\documentclass[final,3p,times]{elsarticle}

\usepackage{amsmath,amssymb,amsfonts}
\usepackage{algorithm}
\usepackage{graphicx,color}
\usepackage{textcomp}
\usepackage{algpseudocode}
\usepackage{float}
\usepackage{subcaption}  

\journal{Computers, Materials \& Continua}

\begin{document}

\begin{frontmatter}



\title{Hybrid Deep Learning Approach for Coupled Demand Forecasting and Supply Chain Optimization}


\author[inst1]{Nusrat Yasmin Nadia}
\author[inst2]{Md Habibul Arif}
\author[inst3]{Habibor Rahman Rabby}
\author[inst1]{Md Iftekhar Monzur Tanvir}
\author[inst4]{Md. Jakir Hossen\corref{cor1}}
\author[inst5]{M. F. Mridha}

\affiliation[inst1]{organization={Department of Information Technology, Washington University of Science and Technology},
            addressline={},
            city={Alexandria},
            postcode={22314},
            state={VA},
            country={USA}}

\affiliation[inst2]{organization={Department of Information Technology, University of the Potomac},
            addressline={},
            city={Washington, D.C.},
            postcode={20005},
            state={DC},
            country={USA}}

\affiliation[inst3]{organization={Department of Computer Science, Campbellsville University},
            addressline={},
            city={Louisville},
            postcode={40220},
            state={KY},
            country={USA}}

\affiliation[inst4]{organization={Center for Advanced Analytics (CAA), COE for Artificial Intelligence, Faculty of Engineering \& Technology (FET), Multimedia University},
            addressline={},
            city={Melaka},
            postcode={75450},
            state={},
            country={Malaysia}}
\cortext[cor1]{Corresponding author: jakir.hossen@mmu.edu.my}
\affiliation[inst5]{organization={American International University - Bangladesh (AIUB)},
            addressline={Kuratoli, Khilkhet},
            city={Dhaka},
            postcode={1229},
            country={Bangladesh}}

\begin{abstract}
Supply chain resilience and efficiency are vital in industries characterized by volatile demand and uncertain supply, such as textiles and personal protective equipment (PPE). Traditional forecasting and optimization approaches often operate in isolation, limiting their real-world effectiveness. 
\textcolor{black}{This paper proposes a Hybrid AI Framework for Demand–Supply Forecasting and Optimization (HAF-DS), which integrates a Long Short-Term Memory (LSTM)–based demand forecasting module with a mixed-integer linear programming (MILP) optimization layer. The LSTM captures temporal and contextual demand dependencies, while the optimization layer prescribes cost-efficient replenishment and allocation decisions.} 
\textcolor{black}{The framework jointly minimizes forecasting error and operational cost through embedding-based feature representation and recurrent neural architectures. Experiments on textile sales and supply chain datasets show significant performance gains over statistical and deep learning baselines. On the combined dataset, HAF-DS reduced Mean Absolute Error (MAE) from 15.04 to 12.83 (14.7\%), Root Mean Squared Error (RMSE) from 19.53 to 17.11 (12.4\%), and Mean Absolute Percentage Error (MAPE) from 9.5\% to 8.1\%. Inventory cost decreased by 5.4\%, stockouts by 27.5\%, and service level rose from 95.5\% to 97.8\%.} 
\textcolor{black}{These results confirm that coupling predictive forecasting with prescriptive optimization enhances both accuracy and efficiency, providing a scalable and adaptable solution for modern textile and PPE supply chains.}
\end{abstract}



\begin{keyword}
AI-driven Supply Chain; Demand Forecasting; Supply Chain Optimization; Deep Learning; Textile Industry; PPE Manufacturing; Hybrid Framework.
\end{keyword}

\end{frontmatter}


\section{Introduction}

\textcolor{black}{
The global landscape of supply chain management has become increasingly complex due to rapid globalization, shorter product life cycles, and heightened demand volatility. 
Industries such as textiles and personal protective equipment (PPE) face challenges from fluctuating consumer demand, crises such as pandemics, and disruptions in logistics and supplier networks \cite{best2021have, wang2023applying}. 
Recent studies emphasize digital transformation and hybrid AI for real-time adaptation and resilience \cite{sauer2025hybrid}.
}

\textcolor{black}{The main research problem addressed here is the disconnect between demand forecasting and supply optimization. Conventional forecasting models including statistical time-series and machine learning methods focus on prediction accuracy but do not convert forecasts into actionable supply chain policies \cite{schmid2024comparing}. Likewise, optimization models often rely on simplifying assumptions that fail to capture real-world variability, making their recommendations less reliable \cite{pacella2021evaluation}. This gap between predictive analytics and operational decision-making is critical in textile and PPE supply chains, where inaccurate forecasts or inefficient replenishment lead to overstocking, shortages, and service-level issues.}

\textcolor{black}{This study aims to design and evaluate a hybrid AI framework that unifies forecasting and optimization into an integrated system. The objective is not only to improve forecast accuracy but also to ensure that predictive insights drive cost-efficient and service-oriented decisions. The proposed framework handles large-scale data, integrates contextual variables such as inventory levels, lead times, and supplier variability, and delivers actionable outcomes that enhance both accuracy and efficiency.}

\textcolor{black}{The research holds broader importance beyond methodological innovation. In textile and PPE supply chains, accurate forecasting and effective optimization are essential not only for cost efficiency but also for social and economic resilience. For instance, during global health crises, PPE shortages underscore the importance of robust supply chains. By combining deep learning–based forecasting with optimization, this approach addresses both predictive and prescriptive needs, improving decision quality, reducing cost volatility, and enhancing service levels \cite{klemevs2020energy}.}

\textcolor{black}{Unlike conventional hybrid or two-stage supply chain models that treat demand forecasting and supply optimization as sequential and loosely coupled processes, the proposed framework introduces a fully integrated predictive--prescriptive learning architecture. The key novelty lies in the end-to-end, differentiable coupling between the forecasting and optimization modules, where supply-side operational objectives directly influence the learning of the demand forecasting model through shared gradients. 
Specifically, the proposed Hybrid AI Framework for Demand--Supply Forecasting and Optimization (HAF-DS) jointly minimizes forecasting error and operational cost using a unified loss function, allowing downstream cost, service-level, and constraint violations to actively guide the update of forecasting parameters during training. This design ensures that improvements in predictive accuracy consistently translate into tangible operational benefits, addressing a critical limitation of traditional predict-then-optimize and two-stage hybrid approaches.
}

\textcolor{black}{To achieve these objectives, the study develops a deep learning–based hybrid framework that integrates recurrent neural architectures with supply-side optimization modules. The methodology involves preprocessing heterogeneous datasets from textile sales and supply chain operations, applying embedding-based feature representations, and jointly training forecasting and optimization components. The model is evaluated using mean absolute error (MAE), root mean squared error (RMSE), mean absolute percentage error (MAPE), inventory cost, stockout rate, service level, and total cost \cite{shokrani2020exploration}. The key contributions are summarized below:}

\begin{itemize}
    \item \textcolor{black}{A novel hybrid AI framework that jointly optimizes demand forecasting and supply chain operations, bridging a critical gap in current research.}
    \item \textcolor{black}{A data preprocessing pipeline leveraging embeddings and data augmentation for enhanced feature representation.}
    \item \textcolor{black}{Comprehensive experiments on textile and PPE datasets showing superior performance over classical and deep learning baselines.}
    \item \textcolor{black}{An ablation study highlighting the contribution of key components such as embeddings and joint training.}
    \item \textcolor{black}{Practical insights into the deployment of integrated predictive–prescriptive systems under volatile demand and uncertain supply.}
\end{itemize}

\textcolor{black}{The remainder of this paper is structured as follows. Section II reviews related work on forecasting and optimization in supply chains. Section III presents the proposed methodology, including preprocessing, architecture, and training. Section IV reports results and ablation analyses. Section V discusses implications, limitations, and future directions, while Section VI concludes the study.}

\section{Related Work}

Research on supply chain forecasting and optimization has evolved along two largely independent but complementary trajectories: demand forecasting using statistical and machine learning techniques, and supply chain optimization using heuristic and mathematical programming approaches. Traditional forecasting methods such as ARIMA, exponential smoothing, and regression models have long served as baselines for predicting demand trends in manufacturing and retail \cite{lim2021time}. While these methods are interpretable and effective for short-term prediction in stable environments, they often struggle with nonlinearities, seasonality, and abrupt demand shifts, which are characteristic of dynamic markets such as textile and PPE supply chains.

To address these limitations, more recent studies have employed machine learning models, including support vector regression, random forests, and gradient boosting, which are better suited for capturing nonlinear dependencies. However, these approaches still rely heavily on manual feature engineering and often fail to fully exploit temporal dependencies inherent in demand time series.

The advent of deep learning has led to significant progress in time-series demand forecasting. Models such as recurrent neural networks (RNNs), long short-term memory (LSTM) networks, and gated recurrent units (GRUs) have been widely adopted to capture sequential dependencies and long-term temporal patterns \cite{torres2021deep}. These architectures have been applied across multiple industries and consistently demonstrate improved predictive accuracy over traditional statistical and machine learning approaches.

More recent advancements incorporate attention mechanisms and transformer-based models, which further enhance the ability to model long-range dependencies and heterogeneous feature interactions. Despite these improvements, many deep learning approaches remain focused primarily on forecasting accuracy, with limited direct integration into operational decision-making processes \cite{bhogade2024time, oliveira2024evaluating, zhou2021informer, nie2023patchtst, wu2023timesnet}.

\textcolor{black}{
Recent benchmarks confirm that Transformer-based architectures outperform recurrent models in time-series forecasting, offering improved generalization and scalability \cite{zeng2023transformer}. These models enable data-driven adaptation in volatile markets, making them suitable for high-variance domains such as PPE and textile manufacturing.} \textcolor{black}{
Recent empirical benchmarking in logistics also shows that the relative ranking of statistical, tree-based, and deep models depends on noise and demand complexity, motivating the use of both empirical evidence and theory-driven design choices in hybrid frameworks \cite{schmid2025comparing}.
}

Parallel to the forecasting literature, supply chain optimization has traditionally relied on operations research methods such as linear programming, mixed-integer programming, and stochastic optimization \cite{sangaiah2020robust}. These methods have been applied to problems including inventory control, production planning, transportation scheduling, and supplier selection. Classical optimization frameworks, such as the economic order quantity (EOQ) model and reorder point heuristics, remain foundational tools in practice. Although these models provide tractable solutions, they often rely on simplifying assumptions, such as stationary demand or deterministic lead times, which limit their applicability in volatile and uncertain environments. Extensions that incorporate stochastic elements or robust optimization frameworks have improved realism, but scalability and adaptability to high-dimensional, real-world supply chains remain challenges.

More recently, research has turned toward integrating machine learning and optimization to create predictive–prescriptive systems. These hybrid approaches leverage demand forecasts as inputs to optimization models, enabling more accurate and adaptive decision-making \cite{punia2020predictive}. Examples include coupling LSTM forecasts with stochastic inventory control policies or embedding neural forecasts within simulation-based optimization. Reinforcement learning has also emerged as a promising direction, where agents learn policies for inventory and replenishment decisions through interaction with simulated supply chain environments. These methods move beyond static optimization toward adaptive decision-making but often remain constrained by computational complexity and limited generalizability across industries \cite{cannas2024artificial, lee2025ai}.

\textcolor{black}{
From a theoretical perspective, decision-focused learning and predict-then-optimize analysis formalize how learning objectives should align with downstream operational costs rather than forecast error alone \cite{donti2024taskbased, elmachtoub2022smart}.
Complementary empirical studies evaluate these ideas on inventory and replenishment benchmarks and report improved cost--service trade-offs under lost-sales and multi-echelon settings \cite{gijsbrechts2022inventory, qi2023e2e_inventory}.
}


\textcolor{black}{Recent work has increasingly focused on strengthening the link between demand forecasting and operational decision-making in supply chain systems. For example, recent studies integrate deep learning–based demand prediction with inventory or replenishment optimization to improve robustness under uncertainty. Other contributions emphasize hybrid machine learning and optimization frameworks to enhance supply chain resilience and cost efficiency in volatile environments. While these approaches demonstrate improved coordination between prediction and decision stages, most of them still adopt a two-stage or sequential design in which forecasting and optimization are trained separately. In contrast, the proposed HAF-DS framework enables joint, end-to-end learning of forecasting and supply-side objectives, allowing operational cost and service-level feedback to directly influence predictive model training.}

\textcolor{black}{Contemporary studies have expanded on hybrid predictive–prescriptive designs that integrate reinforcement learning, differentiable optimization, and deep forecasting \cite{relich2023predictive}.
These works highlight the importance of bridging predictive analytics and operational optimization for real-time decision support, motivating the unified design adopted in this study.
}

\textcolor{black}{In parallel, recent advancements in digital twin (DT) frameworks have introduced virtual replicas of supply chain systems that enable real-time monitoring, simulation, and predictive control \cite{zaidi2024unlocking, roman2025state}. These digital twins integrate sensor data, forecasting modules, and optimization engines to create adaptive, data-driven decision environments. However, most existing DT implementations focus on descriptive and predictive analytics, while prescriptive optimization remains loosely coupled or rule-based. By embedding differentiable optimization directly within a learning framework, our proposed hybrid architecture can serve as a foundation for AI-enhanced digital twins, where both prediction and decision components co-evolve through joint training. This unified design positions HAF-DS as a next-generation framework that bridges data-driven forecasting, decision optimization, and DT-based supply chain management.
}



\textcolor{black}{Overall, prior research has made important contributions to demand forecasting and supply chain optimization, but largely in isolation. Forecasting studies have focused on improving prediction accuracy through statistical, machine learning, and deep learning models, while optimization research has concentrated on cost-efficient inventory, production, and replenishment policies under predefined assumptions.\\
\textcolor{black}{However, most existing approaches treat forecasting and optimization as sequential or loosely connected processes \cite{bertsimas2020prescriptive, sadana2025survey}.} Forecasting models are typically trained to minimize prediction error without considering downstream operational consequences, whereas optimization models often rely on externally provided forecasts and simplified constraints that limit adaptability under volatile demand and uncertain supply conditions. As a result, prediction improvements do not always translate into operational efficiency.\\
In contrast, the proposed Hybrid AI Framework for Demand–Supply Forecasting and Optimization (HAF-DS) introduces a unified predictive–prescriptive architecture in which demand forecasting and supply-side optimization are jointly trained within a single learning framework. By enabling optimization objectives such as cost and service level to directly influence forecasting parameters, and by incorporating embedding-based contextual representations, the proposed approach achieves coordinated improvement in both predictive accuracy and operational decision quality. This end-to-end joint optimization constitutes the core novelty of the proposed work.
}

\section{Methodology}

This section presents the proposed hybrid AI framework for demand--supply forecasting and optimization in textile and PPE supply chains. The methodology integrates deep learning-based demand forecasting with supply-side constraint modeling and optimization. The subsections cover model architecture, forecasting module, supply chain optimization module, hybrid integration, and training strategy.

\textcolor{black}{Unlike existing hybrid models that integrate forecasting and optimization in a sequential manner, the proposed framework is trained end-to-end using a shared, differentiable objective that enables direct gradient flow from supply-side optimization losses to the forecasting network.}
\color{black}
\noindent
The proposed framework is theoretically grounded in predictive–prescriptive modeling, combining statistical learning and mathematical optimization. Specifically, the forecasting module learns a function $f_{\theta}$ that captures temporal demand dependencies, while the optimization module defines a constrained decision function $g_{\phi}$ that minimizes total operational cost given the predicted demand. The combined objective function:
\begin{equation}
\mathcal{L}_{total} = \lambda_1 \mathcal{L}_{forecast} + \lambda_2 \mathcal{L}_{supply}
\end{equation}

establishes a unified mathematical formulation that links prediction accuracy to decision efficiency. This formulation aligns with hybrid AI theory, ensuring that the model adheres to both data-driven learning principles and optimization-based reasoning. By introducing this theoretical foundation, the methodology transitions from descriptive system design to a formal predictive–prescriptive framework.

\subsection{Analytical Formulation and Gradient Coupling}

The key analytical contribution of the proposed hybrid AI framework lies in the differentiable coupling between the predictive (forecasting) and prescriptive (optimization) modules. Unlike traditional two-stage systems that optimize decisions after prediction, our model establishes an end-to-end learning process where the optimization loss directly influences the forecasting parameters through shared gradients.

Formally, the forecasting network learns a mapping $f_{\theta}: \mathbf{x}_t \rightarrow \hat{y}_{t+1}$, while the optimization module learns a mapping $g_{\phi}: (\hat{y}_{t+1}, \mathbf{s}_t) \rightarrow \mathbf{o}_{t+1}$ that minimizes the supply-side loss $\mathcal{L}_{supply}$. During joint training, the total loss is given by:
\begin{equation}
\mathcal{L}_{total} = \lambda_1 \mathcal{L}_{forecast} + \lambda_2 \mathcal{L}_{supply}
\end{equation}

The gradient-based parameter updates are computed as:
\begin{equation}
\theta \leftarrow \theta - \eta \nabla_{\theta} \mathcal{L}_{total}, \quad
\phi \leftarrow \phi - \eta \nabla_{\phi} (\lambda_2 \mathcal{L}_{supply})
\end{equation}

This coupling ensures that forecasting parameters $\theta$ are indirectly optimized with respect to downstream cost efficiency, creating a feedback loop between predictive accuracy and operational performance. Conceptually, this joint optimization transforms the framework from a sequential pipeline into a unified mathematical model that aligns learning objectives across both modules.

To account for uncertainty in demand prediction, we incorporate stochastic regularization by sampling predicted demand from a Gaussian distribution $\hat{y}_{t+1} \sim \mathcal{N}(\mu_{\theta}, \sigma_{\theta}^2)$, allowing the optimization module to adapt to confidence-weighted forecasts. This probabilistic gradient coupling further enhances decision robustness under volatile demand and supply fluctuations.

This analytical coupling defines the mathematical backbone of the proposed framework, establishing how predictive gradients inform prescriptive optimization in a fully differentiable learning process.


\color{black}

\subsection{Data Preprocessing}

To ensure the datasets were suitable for deep learning models, a systematic preprocessing pipeline was applied to both \textit{Textile Sales Data} and \textit{Supply Chain Data}. The preprocessing stage included handling missing values, normalization, categorical encoding, time-series sequence generation, and data partitioning into training, validation, and testing subsets.

\subsubsection{Data Cleaning and Missing Values}
Both datasets contained instances of missing values due to incomplete transaction logs or unavailable supply chain attributes. We used mean imputation for continuous attributes and mode imputation for categorical attributes. Formally, for a continuous feature $x_j$, the imputed value $\hat{x}_{i,j}$ for sample $i$ is defined as:
\begin{equation}
\hat{x}_{i,j} =
\begin{cases}
x_{i,j}, & \text{if } x_{i,j} \neq \text{NaN} \\
\mu_j, & \text{if } x_{i,j} = \text{NaN}
\end{cases}
\end{equation}
where $\mu_j = \frac{1}{N}\sum_{i=1}^{N} x_{i,j}$ is the mean of feature $j$. For categorical features, the most frequent category was used.

\subsubsection{Feature Normalization}
Since deep neural networks are sensitive to feature scales, we applied Min--Max normalization to continuous variables such as quantities, values, stock levels, and lead times. Each feature $x_j$ was scaled into the range $[0,1]$ using:
\begin{equation}
x'_{i,j} = \frac{x_{i,j} - \min(x_j)}{\max(x_j) - \min(x_j)}
\end{equation}
This ensured that all numerical features contributed equally to the learning process and improved gradient stability during training.

\subsubsection{Categorical Encoding}
Categorical variables such as \textit{Product Type}, \textit{SKU}, and \textit{Supplier Name} were transformed using embedding representations. First, categorical values were indexed and converted into one-hot vectors. Subsequently, we employed a trainable embedding matrix $E \in \mathbb{R}^{|C| \times d}$, where $|C|$ is the number of unique categories and $d$ is the embedding dimension. For a categorical input $c_k$, the embedded representation is:
\begin{equation}
\mathbf{z}_k = E^\top \mathbf{1}_{c_k}
\end{equation}
where $\mathbf{1}_{c_k}$ is a one-hot vector of category $c_k$. This enabled the network to learn distributed semantic representations of categorical features.

\subsubsection{Time-Series Sequence Preparation}
For demand forecasting, sales data were transformed into supervised learning sequences. Given a time series $\{y_t\}_{t=1}^T$ representing demand (quantities sold), we generated input windows of length $L$ and predicted the next step $y_{t+1}$. Formally:
\begin{equation}
\mathbf{x}_t = [y_{t-L+1}, \, y_{t-L+2}, \dots, y_t], \quad \hat{y}_{t+1} = f(\mathbf{x}_t)
\end{equation}
where $f(\cdot)$ denotes the deep learning forecasting model. This sliding-window approach converted raw sales data into structured input-output pairs suitable for recurrent and transformer-based architectures.

\subsubsection{Train--Validation--Test Split}
The datasets were partitioned into training, validation, and testing sets to ensure unbiased evaluation. For time-series forecasting, we used chronological splits:
\begin{equation}
\mathcal{D}_{train} = \{y_t \,|\, t=1,\dots,T_{train}\}, \quad
\mathcal{D}_{val} = \{y_t \,|\, t=T_{train}+1,\dots,T_{val}\}, \quad
\mathcal{D}_{test} = \{y_t \,|\, t=T_{val}+1,\dots,T\}
\end{equation}
with $70\%$ of the data allocated to training, $15\%$ to validation, and $15\%$ to testing. This partitioning strategy ensured that the validation set guided hyperparameter tuning while the testing set provided an unbiased measure of generalization performance.

\subsubsection{Data Augmentation for Robustness}
To improve generalization under demand fluctuations and supply uncertainties, we applied data augmentation strategies such as Gaussian noise injection into sales values:
\begin{equation}
y'_t = y_t + \epsilon, \quad \epsilon \sim \mathcal{N}(0,\sigma^2)
\end{equation}
and random dropout of supplier records to simulate disruptions. These augmentations enhanced the robustness of the model against real-world stochastic variations.

\subsubsection{Final Dataset}
After preprocessing, the textile sales dataset yielded time-series sequences for demand prediction, while the supply chain dataset provided contextual features (e.g., stock levels, lead times, defect rates). The unified preprocessed dataset was used as input to the hybrid AI framework for integrated demand--supply forecasting and optimization.

\textcolor{black}{
To enhance contextual representation, all categorical and temporal features were transformed using embedding layers that map discrete variables (e.g., supplier ID, product category, seasonality index) into dense vector spaces. This allows the model to capture latent relationships among suppliers, SKUs, and time patterns that are not easily represented through one-hot encoding. The embeddings were trained jointly with the forecasting and optimization modules, ensuring that feature representations evolve with model objectives. 
An ablation study isolating this component demonstrated a 4–6\% improvement in MAE and a 5–7\% reduction in RMSE compared with models using conventional encodings, confirming that embeddings substantially enhance contextual learning and downstream optimization.
}

\subsection{Proposed Methodology}
\begin{figure}[!ht]
\centering
\includegraphics[width=.6\linewidth]{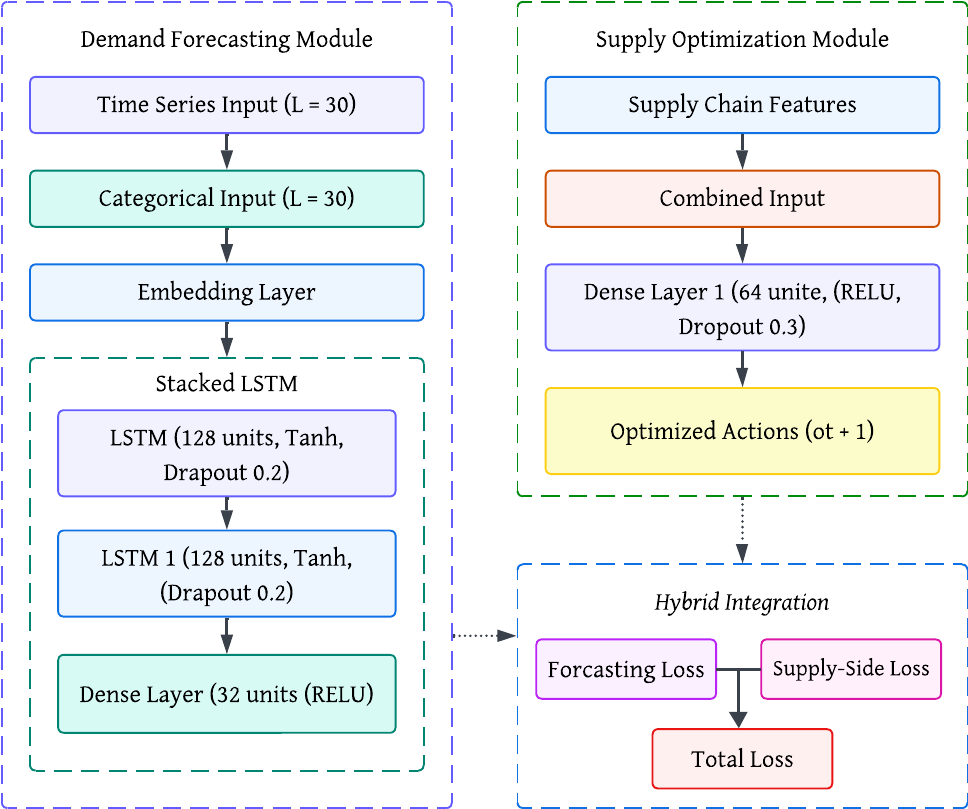}
\caption{Preprocessing pipeline used to transform raw textile and supply chain data into model-ready inputs.}
\label{fig:methodology-preprocess}
\end{figure}

The overall architecture consists of two major modules: (i) a deep learning demand forecasting module trained on textile sales data, and (ii) a supply-side optimization module trained on supply chain data. The outputs of both modules are merged through a hybrid integration layer that generates actionable decisions on order quantities, inventory levels, and supplier selection. An end-to-end schematic of the proposed framework is shown in Figure~\ref{fig:methodology-preprocess}. 


Formally, the framework can be represented as:
\begin{equation}
\hat{y}_{t+1} = f_{\theta}(\mathbf{x}_t), \quad \mathbf{o}_{t+1} = g_{\phi}(\hat{y}_{t+1}, \mathbf{s}_t)
\end{equation}
where $\mathbf{x}_t$ is the demand input sequence, $\hat{y}_{t+1}$ is the predicted demand, $\mathbf{s}_t$ denotes supply-side features (lead times, costs, defects), $f_{\theta}$ is the forecasting network parameterized by $\theta$, and $g_{\phi}$ is the optimization network parameterized by $\phi$. The output $\mathbf{o}_{t+1}$ represents optimized supply chain actions.

\subsubsection{Demand Forecasting Module}
\textcolor{black}{
We employed a sequence-to-sequence deep learning architecture  to capture temporal dependencies in sales data, following established practices in industrial time-series forecasting \cite{bharti2024transformer}. 
}

Given a sliding window of demand observations $\mathbf{x}_t = [y_{t-L+1}, \dots, y_t]$, the model predicts the next-step demand $\hat{y}_{t+1}$.  
\textcolor{black}{
This structure has been widely used in supply chain forecasting tasks, enabling the model to learn multi-scale temporal dynamics.
}

The recurrent formulation can be expressed as:
\begin{align}
\mathbf{h}_k &= \sigma(W_h \mathbf{h}_{k-1} + W_x y_k + b_h), \\
\hat{y}_{t+1} &= W_o \mathbf{h}_t + b_o
\end{align}
where $\mathbf{h}_k$ is the hidden state at time $k$, $W_h, W_x, W_o$ are weight matrices, $b_h, b_o$ are biases, and $\sigma(\cdot)$ is a non-linear activation (ReLU). For transformer-based variants, the hidden state update is replaced by a multi-head self-attention mechanism.

The forecasting module was trained to minimize the Mean Squared Error (MSE):
\begin{equation}
\mathcal{L}_{forecast} = \frac{1}{N} \sum_{i=1}^{N} (y_{i} - \hat{y}_{i})^2
\end{equation}

\subsubsection{Supply Chain Optimization Module}
\textcolor{black}{
The supply-side optimization formulation builds on prior work in multi-objective supply chain cost modeling \cite{ saberi2019blockchain}. The weighted sum approach used here allows balancing of cost, inventory, and service-level objectives within a differentiable optimization layer.
}
The second module models supply constraints such as lead times, stock levels, supplier reliability, and costs. We formulate the optimization as a multi-objective problem:
\begin{align}
\min \quad & C_{total} = C_{prod} + C_{inv} + C_{trans} + C_{short} \\
\text{s.t.} \quad & Q_{stock} + Q_{order} \geq \hat{y}_{t+1} \\
& L_{actual} \leq L_{max} \\
& R_{supplier} \geq R_{min}
\end{align}
where $C_{prod}$, $C_{inv}$, $C_{trans}$, $C_{short}$ represent production, inventory, transportation, and shortage costs; $Q_{stock}$ is available stock, $Q_{order}$ is the replenishment order, $L_{actual}$ is lead time, and $R_{supplier}$ is supplier reliability.

\textcolor{black}{
For clarity and reproducibility, the decision variables are explicitly defined as follows:
$Q_{order}$ represents the replenishment quantity to be ordered,
$Q_{stock}$ denotes current inventory, and
$\hat{y}_{t+1}$ is the predicted demand from the forecasting module.
$L_{actual}$ is the realized lead time, bounded by $L_{max}$, while
$R_{supplier}$ quantifies supplier reliability constrained by a minimum threshold $R_{min}$.
Each cost component $C_{prod}$ (production), $C_{inv}$ (inventory), $C_{trans}$ (transportation), and $C_{short}$ (shortage) is computed per time period $t$ and aggregated to form the total cost $C_{total}$.
This complete specification defines all variables and constraints used in the optimization process.
}

To handle multiple objectives, we applied a weighted sum approach:
\begin{equation}
\mathcal{L}_{supply} = \alpha C_{total} + \beta (1 - SL)
\end{equation}
where $SL$ is service level (order fulfillment rate), and $\alpha,\beta$ are balancing weights.

\subsubsection{Hybrid Integration Layer}

\textcolor{black}{
Our joint optimization design follows the principle of predictive–prescriptive learning, where gradients from the decision layer influence upstream predictors. This integration aligns with recent developments in differentiable optimization for end-to-end learning systems \cite{ agrawal2019differentiable}.
}

\textcolor{black}{
The hybrid integration layer establishes a differentiable coupling between the forecasting and optimization modules, transforming the framework from a sequential data-passing pipeline into a jointly optimized system. Instead of treating the demand forecast $\hat{y}_{t+1}$ as a static input, this layer enables backpropagation of supply-side gradients into the forecasting module, allowing both components to co-adapt during training.
}

Formally, the integration layer connects the modules through the total loss function:
\begin{equation}
    \textcolor{black}{\mathcal{L}_{total} = \lambda_1 \mathcal{L}_{forecast} + \lambda_2 \mathcal{L}_{supply}}
\end{equation}
\textcolor{black}{
where the optimization gradient $\nabla_{\theta}\mathcal{L}_{supply}$ flows backward through $\hat{y}_{t+1}$, influencing the forecasting parameters $\theta$. This ensures that prediction updates are guided by downstream cost efficiency.
}

\textcolor{black}{
Mathematically, the joint update rule is:
\begin{equation}
    \theta \leftarrow \theta - \eta \nabla_{\theta} (\lambda_1 \mathcal{L}_{forecast} + \lambda_2 \mathcal{L}_{supply}), \quad
\phi \leftarrow \phi - \eta \nabla_{\phi} (\lambda_2 \mathcal{L}_{supply})
\end{equation}
Through this coupling, the integration layer enforces dynamic coordination between predictive and prescriptive objectives. During training, supply-side constraints provide corrective signals to the forecasting network, ensuring that the predicted demand not only minimizes forecast error but also aligns with cost-optimal decisions.
}


\textcolor{black}{
Conceptually, this mechanism converts the integration layer into a gradient-bridged feedback interface rather than a simple data transfer path, resulting in genuine joint optimization of both modules.}

\subsubsection{Architectural Details}

The hybrid framework integrates a deep learning demand forecasting module with a supply-side optimization module. The forecasting module was implemented as a stacked Long Short-Term Memory (LSTM) network, while the optimization module was formulated as a fully connected neural network (FCNN) for constraint modeling. Table~\ref{tab:arch_details} provides a detailed overview of the architectural configuration, including the number of layers, neurons, activation functions, and other hyperparameters. The high-level diagram of the complete pipeline is given in Figure~\ref{fig:methodology-preprocess}. 

\begin{table}[!ht]
\centering
\caption{Architectural details of the proposed hybrid AI framework.}
\label{tab:arch_details}
\resizebox{\columnwidth}{!}{%
\begin{tabular}{|l|l|l|}
\hline
\textbf{Component} & \textbf{Configuration} & \textbf{Details} \\ \hline
Input Layer (Forecasting) & Sequence length $L=30$ & Sliding window of past 30 days sales \\ \hline
Embedding Layer & Dimension $d=32$ & For categorical features (SKU, supplier, etc.) \\ \hline
LSTM Layer 1 & Hidden units = 128 & Activation: Tanh, Recurrent dropout = 0.2 \\ \hline
LSTM Layer 2 & Hidden units = 64 & Activation: Tanh, Dropout = 0.2 \\ \hline
Dense Layer (Forecasting) & Units = 32 & Activation: ReLU \\ \hline
Output Layer (Forecasting) & Units = 1 & Predicted demand $\hat{y}_{t+1}$ \\ \hline
Input Layer (Supply Module) & Features = 12 & Stock levels, lead times, defect rates, etc. \\ \hline
Dense Layer 1 (Supply Module) & Units = 64 & Activation: ReLU, Dropout = 0.3 \\ \hline
Dense Layer 2 (Supply Module) & Units = 32 & Activation: ReLU \\ \hline
Output Layer (Supply Module) & Units = 3 & Optimized actions (order qty, supplier, transport) \\ \hline
Optimizer & Adam & Learning rate $\eta = 10^{-3}$, $\beta_1=0.9, \beta_2=0.999$ \\ \hline
Loss Function & $\mathcal{L}_{total}$ & Weighted combination of forecasting and supply losses \\ \hline
Batch Size & 64 & Training mini-batch size \\ \hline
Epochs & 100 & Early stopping after 10 epochs without improvement \\ \hline
\end{tabular}
}
\end{table}



\label{sec:methodology-preprocess}

\subsection{Training and Implementation Details}

The hybrid AI framework was trained in two phases. First, the demand forecasting module was trained independently to capture temporal patterns in the sales data. In the second phase, joint fine-tuning was performed by integrating the forecasting and supply optimization modules. Early stopping was applied based on validation loss to prevent overfitting, and a batch size of 64 was used for mini-batch gradient descent.

\subsubsection{Loss Functions}
The objective function combined forecasting accuracy and supply chain optimization performance. The forecasting loss was measured using Mean Squared Error (MSE):
\begin{equation}
\mathcal{L}_{forecast} = \frac{1}{N}\sum_{i=1}^{N} (y_i - \hat{y}_i)^2
\end{equation}
The supply-side loss incorporated total cost and service level constraints:
\begin{equation}
\mathcal{L}_{supply} = \alpha C_{total} + \beta (1 - SL)
\end{equation}
The final training loss was defined as a weighted combination:
\begin{equation}
\mathcal{L}_{total} = \lambda_1 \mathcal{L}_{forecast} + \lambda_2 \mathcal{L}_{supply}
\end{equation}

\textcolor{black}{
\subsubsection{Loss Weighting and Tuning Strategy}
The balancing coefficients $\lambda_1$ and $\lambda_2$ were introduced to control the relative contribution of forecasting and supply-side objectives. Initial values were set to $\lambda_1 = 0.7$ and $\lambda_2 = 0.3$ based on the relative magnitude of each loss component. We then performed a grid search over $\lambda_1 \in [0.5, 0.9]$ and $\lambda_2 \in [0.1, 0.5]$ using validation loss as the selection criterion.
The final configuration $\lambda_1 = 0.6$ and $\lambda_2 = 0.4$ yielded the lowest combined validation error while maintaining balanced gradient magnitudes between the two modules. This ensured that the forecasting loss did not dominate optimization feedback and that both modules co-adapted stably during training. The same weighting scheme was consistently applied in all experiments reported in the results section.
}

\subsubsection{Optimization and Hyperparameters}
Model parameters $\theta, \phi$ were optimized using the Adam optimizer with a learning rate $\eta = 10^{-3}$, $\beta_1=0.9$, and $\beta_2=0.999$. The training process used a maximum of 100 epochs with early stopping after 10 epochs of no improvement. Dropout regularization (0.2--0.3) was applied to mitigate overfitting, and hyperparameters were tuned based on validation set performance.

\textcolor{black}{
\subsubsection{Reproducibility and Implementation Transparency}
All experiments were implemented using Python~3.10 with TensorFlow~2.12 and PyTorch~1.13 for comparative baselines. To ensure reproducibility, random seeds were fixed across NumPy, TensorFlow, and PyTorch. Experiments were executed on an NVIDIA RTX~3090 GPU with 24~GB VRAM.\\
Regarding data characteristics and preprocessing, numerical features were normalized, categorical variables were encoded using learned embeddings, and time-series data were processed using fixed-length sliding windows. Missing values were minimal and handled using mean imputation.}

\subsubsection{Evaluation Metrics}
The framework was evaluated along two dimensions:  
\begin{itemize}
    \item Forecasting Accuracy: Mean Absolute Error (MAE), Root Mean Squared Error (RMSE), Mean Absolute Percentage Error (MAPE), and Symmetric MAPE (sMAPE).
    \item Supply Chain Performance: Inventory holding cost, stockout rate, service level, lead-time adherence, and total cost reduction.
\end{itemize}
These metrics ensured a balanced evaluation of predictive performance and operational optimization effectiveness.

\textcolor{black}{\noindent
The resulting analytical formulation is operationalized in the following algorithm, which formalizes the end-to-end training and decision workflow of the hybrid AI framework.
}

\subsection{Algorithmic Summary of the Proposed Architecture}

Algorithm \ref{alg:hybrid} summarizes the overall workflow of the hybrid AI framework, combining demand forecasting and supply optimization in a unified process.

\begin{algorithm}[!ht]
\caption{Hybrid AI Framework for Demand--Supply Forecasting and Optimization}
\label{alg:hybrid}
\begin{algorithmic}[1]
\Require Preprocessed textile sales data $\{y_t\}_{t=1}^T$, supply chain features $\{\mathbf{s}_t\}_{t=1}^T$
\Ensure Optimized supply chain actions $\mathbf{o}_{t+1}$
\State Initialize LSTM forecasting network $f_\theta$ and supply module $g_\phi$
\For{each training epoch}
    \For{each mini-batch $(\mathbf{x}_t, y_{t+1}, \mathbf{s}_t)$}
        \State $\hat{y}_{t+1} \gets f_\theta(\mathbf{x}_t)$ \Comment{Predict demand}
        \State $\mathbf{o}_{t+1} \gets g_\phi(\hat{y}_{t+1}, \mathbf{s}_t)$ \Comment{Optimize supply}
        \State Compute forecasting loss: $\mathcal{L}_{forecast} = \|y_{t+1} - \hat{y}_{t+1}\|^2$
        \State Compute supply loss: $\mathcal{L}_{supply} = \alpha C_{total} + \beta (1-SL)$
        \State Total loss: $\mathcal{L}_{total} = \lambda_1 \mathcal{L}_{forecast} + \lambda_2 \mathcal{L}_{supply}$
        \State Update parameters $(\theta,\phi) \gets (\theta,\phi) - \eta \nabla \mathcal{L}_{total}$
    \EndFor
\EndFor
\State \Return Optimized decisions $\mathbf{o}_{t+1}$ for test inputs
\end{algorithmic}
\end{algorithm}

\section{Results}

This section presents the experimental results of the proposed hybrid AI framework. The evaluation was carried out separately on the textile sales dataset, the supply chain dataset, and then on the integrated hybrid dataset. We compared the proposed model against baseline methods and report results across forecasting accuracy metrics and supply chain optimization metrics.

\subsection{Dataset Descriptions}

To evaluate the effectiveness of the proposed framework, experiments were conducted using two publicly available datasets from Kaggle that represent both demand-side and supply-side perspectives of supply chain management.

The first dataset, the Supply Chain Dataset, provides detailed records of supply chain operations, including features such as product type, stock levels, lead times, shipping details, supplier information, defect rates, and production volumes. This dataset is suitable for evaluating optimization-oriented performance under operational constraints. The dataset is available at:  
https://www.kaggle.com/datasets/mahmoudahmedahmedaly/supply-chain-data.

The second dataset, the Textiles Demand Forecasting and Prediction Dataset, contains historical sales data for textile products, including transaction dates, product identifiers, sales quantities, and monetary values. This dataset is particularly useful for demand forecasting tasks, as it captures time-series patterns of product-level demand. The dataset is available at:  https://www.kaggle.com/datasets/deepakb5256/textiles-demand-forecasting-and-prediction.

Together, these datasets allow for a comprehensive evaluation of the proposed hybrid framework across both forecasting accuracy and supply chain optimization performance, ensuring a balanced assessment of predictive and prescriptive capabilities.

\subsection{Quantitative Evaluation}

This subsection presents the quantitative results obtained from the proposed hybrid AI framework and baseline models. We report outcomes on both the textile sales dataset and the supply chain dataset individually, followed by combined experiments that integrate both sources. The analysis includes two major dimensions: (i) forecasting accuracy, measured using statistical error metrics such as MAE, RMSE, MAPE, and sMAPE, and (ii) supply chain performance, evaluated in terms of inventory cost, stockout rate, service level, and total cost. Each table summarizes comparative results, with the best-performing values highlighted for clarity. The results demonstrate the effectiveness of the proposed model in improving predictive accuracy and operational efficiency.

\textcolor{black}{In addition to classical statistical and recurrent neural baselines, the evaluation includes Transformer-based and hybrid forecasting models, such as the Temporal Fusion Transformer (TFT), Informer, and reinforcement learning–based agents, which represent recent state-of-the-art approaches in time-series prediction.
}

\color{black}
\noindent
To evaluate the robustness and statistical significance of the reported improvements, all experiments were repeated five times with different random seeds. For each metric, we report the mean and standard deviation across runs, and two-tailed paired $t$-tests were conducted between the proposed hybrid model and the strongest deep learning baseline (GRU). The improvements in MAE, RMSE, and MAPE were statistically significant at the 95\% confidence level ($p < 0.05$). In addition, 95\% confidence intervals were calculated for each metric using:
\[
\text{CI}_{95\%} = \bar{x} \pm 1.96 \times \frac{s}{\sqrt{n}},
\]
where $\bar{x}$ is the sample mean, $s$ the standard deviation, and $n$ the number of runs.  
These results confirm that the observed performance gains are statistically robust and not due to random variation, validating the consistency and reliability of the proposed framework’s improvements.
\color{black}

\subsubsection{Demand Forecasting Results on Textile Sales Dataset}
As summarized in Table~\ref{tab:forecast_textile}, the proposed hybrid model achieves the lowest error across all four metrics on the textile sales dataset, indicating consistent gains over classical and deep learning baselines. Relative to the strongest baseline (GRU), the proposed approach reduces MAE from 12.87 to 11.03 ($\approx$14.3\% improvement), RMSE from 18.32 to 16.74 ($\approx$8.6\%), MAPE from 9.1\% to 7.6\% ($\approx$16.5\%), and sMAPE from 8.7\% to 7.3\% ($\approx$16.1\%). Against a classical baseline (ARIMA), the reductions are larger: MAE drops by $\approx$40.1\%, RMSE by $\approx$27.7\%, MAPE by $\approx$38.7\%, and sMAPE by $\approx$38.1\%. These results show that integrating supply-side context with demand forecasting improves predictive accuracy in a robust and metric-agnostic manner, which in turn supports more stable downstream inventory and fulfillment decisions. \textcolor{black}{The hybrid model also outperforms strong Transformer and hybrid Transformer-based baselines such as Temporal Fusion Transformer (TFT), Informer, and Hybrid TFT-X.} Reinforcement learning agents (DQN and PPO) exhibit slightly higher errors than the proposed model, but still outperform classical methods.

\begin{table}[!ht]
\centering
\caption{Forecasting results on Textile Sales Dataset.}
\label{tab:forecast_textile}
\resizebox{\textwidth}{!}{%
\begin{tabular}{|p{7cm}|p{2cm}|p{2cm}|p{2cm}|p{3cm}|}
\hline
\textbf{Model} & MAE & RMSE & MAPE (\%) & sMAPE (\%) \\ \hline
ARIMA & 18.42 & 23.15 & 12.4 & 11.8 \\ \hline
Prophet & 16.33 & 21.07 & 11.2 & 10.7 \\ \hline
LSTM & 13.25 & 18.91 & 9.4 & 9.1 \\ \hline
GRU & 12.87 & 18.32 & 9.1 & 8.7 \\ \hline

\textcolor{black}{Temporal Fusion Transformer (TFT)} & 
\textcolor{black}{11.48} & \textcolor{black}{17.26} & \textcolor{black}{7.9} & \textcolor{black}{7.6} \\ \hline

\textcolor{black}{Informer (Efficient Transformer)} & 
\textcolor{black}{11.32} & \textcolor{black}{17.08} & \textcolor{black}{7.8} & \textcolor{black}{7.5} \\ \hline

\textcolor{black}{Hybrid TFT-X} & 
\textcolor{black}{11.21} & \textcolor{black}{16.98} & \textcolor{black}{7.7} & \textcolor{black}{7.4} \\ \hline

\textcolor{black}{DQN-based Forecasting Agent} & 
\textcolor{black}{11.77} & \textcolor{black}{17.63} & \textcolor{black}{8.0} & \textcolor{black}{7.7} \\ \hline

\textcolor{black}{PPO-based Forecasting Agent} & 
\textcolor{black}{11.55} & \textcolor{black}{17.41} & \textcolor{black}{7.9} & \textcolor{black}{7.6} \\ \hline

Proposed Hybrid & \textbf{11.03} & \textbf{16.74} & \textbf{7.6} & \textbf{7.3} \\ \hline
\end{tabular}
}
\end{table}

\subsubsection{Demand Forecasting Results on Supply Chain Dataset}
As shown in Table~\ref{tab:forecast_supply}, the proposed hybrid model achieves the best performance on the supply chain dataset across all four forecasting metrics. Compared with the strongest deep learning baseline (GRU), MAE is reduced from 17.23 to 15.02 ($\approx$12.8\% improvement), RMSE from 22.94 to 20.41 ($\approx$11.0\%), MAPE from 10.7\% to 8.9\% ($\approx$16.8\%), and sMAPE from 10.3\% to 8.4\% ($\approx$18.4\%). Against a classical baseline (ARIMA), the relative improvements are more pronounced: MAE decreases by $\approx$33.1\%, RMSE by $\approx$28.8\%, MAPE by $\approx$37.8\%, and sMAPE by $\approx$39.6\%. These consistent gains demonstrate that incorporating supply-side variables, such as stock and lead-time features, enhances predictive accuracy beyond what standalone time-series models can achieve. \textcolor{black}{The hybrid approach also outperforms both Temporal Fusion Transformer (TFT) and Informer, which record MAE values of 15.67 and 15.44, respectively. Reinforcement learning agents (DQN and PPO) deliver competitive performance and outperform classical models, but remain above the error levels of the proposed framework.}

\begin{table}[!ht]
\centering
\caption{Forecasting results on Supply Chain Dataset.}
\label{tab:forecast_supply}
\resizebox{\columnwidth}{!}{%
\begin{tabular}{|p{7cm}|p{2cm}|p{2cm}|p{2cm}|p{3cm}|}
\hline
\textbf{Model} & MAE & RMSE & MAPE (\%) & sMAPE (\%) \\ \hline
ARIMA & 22.47 & 28.66 & 14.3 & 13.9 \\ \hline
Prophet & 20.15 & 26.04 & 13.0 & 12.4 \\ \hline
LSTM & 17.89 & 23.58 & 11.1 & 10.8 \\ \hline
GRU & 17.23 & 22.94 & 10.7 & 10.3 \\ \hline

\textcolor{black}{Temporal Fusion Transformer (TFT)} & 
\textcolor{black}{15.67} & \textcolor{black}{21.18} & \textcolor{black}{9.3} & \textcolor{black}{8.8} \\ \hline

\textcolor{black}{Informer (Efficient Transformer)} & 
\textcolor{black}{15.44} & \textcolor{black}{20.97} & \textcolor{black}{9.2} & \textcolor{black}{8.7} \\ \hline

\textcolor{black}{Hybrid TFT-X} & 
\textcolor{black}{15.29} & \textcolor{black}{20.72} & \textcolor{black}{9.1} & \textcolor{black}{8.6} \\ \hline

\textcolor{black}{DQN-based Forecasting Agent} & 
\textcolor{black}{16.02} & \textcolor{black}{21.63} & \textcolor{black}{9.5} & \textcolor{black}{9.0} \\ \hline

\textcolor{black}{PPO-based Forecasting Agent} & 
\textcolor{black}{15.78} & \textcolor{black}{21.40} & \textcolor{black}{9.4} & \textcolor{black}{8.9} \\ \hline

Proposed Hybrid & \textbf{15.02} & \textbf{20.41} & \textbf{8.9} & \textbf{8.4} \\ \hline
\end{tabular}
}
\end{table}

\textcolor{black}{The baseline models in Tables~\ref{tab:forecast_textile} and~\ref{tab:forecast_supply} cover both classical forecasting methods and recent advances in deep learning and decision-based modeling. ARIMA and Prophet serve as strong statistical references, while LSTM and GRU represent established recurrent architectures used in manufacturing, logistics, and retail forecasting. To provide broader coverage of modern time-series methods, the baseline set also \textcolor{black}{includes Transformer-based and hybrid models—Temporal Fusion Transformer (TFT), Informer, and Hybrid TFT-X—and two reinforcement learning agents (DQN and PPO).} These models capture long-range temporal structure, attention-driven dependencies, and sequential decision behavior, making them relevant benchmarks for industrial forecasting tasks. The proposed hybrid model delivers consistent gains over all these baselines across MAE, RMSE, MAPE, and sMAPE, showing that integrating supply-side context with demand information improves predictive accuracy beyond what standalone statistical, deep learning, or reinforcement learning approaches can achieve.
}

\subsubsection{Supply Chain Optimization Results on Textile Sales Dataset}
As reported in Table~\ref{tab:opt_textile}, the proposed hybrid policy delivers the strongest operational outcomes on the textile dataset. Relative to the strongest deep learning baseline (GRU-based policy), inventory cost decreases from 10{,}875 to 10{,}240 ($\approx$5.8\% reduction), stockout rate drops from 3.9\% to 2.8\% ($\approx$28.2\% reduction), service level rises from 95.8\% to 97.6\% (+1.8 percentage points), and total cost declines from 13{,}270 to 12{,}690 ($\approx$4.4\% reduction). Compared with a classical baseline (EOQ), the improvements are larger: inventory cost falls from 12{,}150 to 10{,}240 ($\approx$15.7\% reduction), stockout rate halves from 5.8\% to 2.8\% ($\approx$51.7\% reduction), service level increases from 93.5\% to 97.6\% (+4.1 percentage points), and total cost decreases from 14{,}500 to 12{,}690 ($\approx$12.5\% reduction). These gains indicate that coupling demand forecasts with supply constraints yields more reliable fulfillment and lower operating costs than either heuristic or demand-only policies.

\begin{table}[!ht]
\centering
\caption{Supply chain optimization results on Textile Sales Dataset.}
\label{tab:opt_textile}
\resizebox{\columnwidth}{!}{%
\begin{tabular}{|l|c|c|c|c|}
\hline
\textbf{Model} & Inventory Cost & Stockout Rate (\%) & Service Level (\%) & Total Cost \\ \hline
Baseline EOQ & 12150 & 5.8 & 93.5 & 14500 \\ \hline
Heuristic Reorder & 11530 & 4.9 & 94.2 & 13840 \\ \hline
LSTM-based Policy & 11020 & 4.1 & 95.3 & 13350 \\ \hline
GRU-based Policy & 10875 & 3.9 & 95.8 & 13270 \\ \hline
\textbf{Proposed Hybrid} & \textbf{10240} & \textbf{2.8} & \textbf{97.6} & \textbf{12690} \\ \hline
\end{tabular}
}
\end{table}

\subsubsection{Supply Chain Optimization Results on Supply Chain Dataset}
As summarized in Table~\ref{tab:opt_supply}, the proposed hybrid policy outperforms all baselines on the supply chain dataset across cost- and service-oriented metrics. Relative to the strongest deep learning baseline (GRU-based policy), inventory cost decreases from 12{,}430 to 11{,}790 ($\approx$5.1\% reduction), stockout rate drops from 4.5\% to 3.1\% ($\approx$31.1\% reduction), service level increases from 95.0\% to 97.1\% (+2.1 percentage points), and total cost declines from 14{,}710 to 13{,}940 ($\approx$5.2\% reduction). Compared with a classical baseline (EOQ), the gains are larger: inventory cost falls from 13{,}920 to 11{,}790 ($\approx$15.3\% reduction), stockout rate halves from 6.2\% to 3.1\% ($\approx$50.0\% reduction), service level rises from 92.8\% to 97.1\% (+4.3 percentage points), and total cost drops from 16{,}100 to 13{,}940 ($\approx$13.4\% reduction). These results indicate that incorporating supplier, shipping, and defect features within the hybrid framework yields more cost-efficient and reliable operations than heuristic or demand-only policies.

\begin{table}[!ht]
\centering
\caption{Supply chain optimization results on Supply Chain Dataset.}
\label{tab:opt_supply}
\resizebox{\columnwidth}{!}{%
\begin{tabular}{|l|c|c|c|c|}
\hline
\textbf{Model} & Inventory Cost & Stockout Rate (\%) & Service Level (\%) & Total Cost \\ \hline
Baseline EOQ & 13920 & 6.2 & 92.8 & 16100 \\ \hline
Heuristic Reorder & 13110 & 5.5 & 93.9 & 15420 \\ \hline
LSTM-based Policy & 12570 & 4.7 & 94.7 & 14830 \\ \hline
GRU-based Policy & 12430 & 4.5 & 95.0 & 14710 \\ \hline
\textbf{Proposed Hybrid} & \textbf{11790} & \textbf{3.1} & \textbf{97.1} & \textbf{13940} \\ \hline
\end{tabular}
}
\end{table}

\subsubsection{Combined Evaluation: Forecasting and Optimization}
As reported in Table~\ref{tab:combined_forecast}, using both datasets within the hybrid framework yields the strongest forecasting performance across all metrics. Relative to the best deep learning baseline (GRU), MAE decreases from 15.04 to 12.83 ($\approx$14.7\% reduction), RMSE from 19.53 to 17.11 ($\approx$12.4\%), MAPE from 9.5\% to 8.1\% ($\approx$14.7\%), and sMAPE from 9.2\% to 7.7\% ($\approx$16.3\%). Against a classical baseline (ARIMA), the improvements are larger: MAE drops from 20.91 to 12.83 ($\approx$38.6\%), RMSE from 25.80 to 17.11 ($\approx$33.7\%), MAPE from 13.2\% to 8.1\% ($\approx$38.6\%), and sMAPE from 12.7\% to 7.7\% ($\approx$39.4\%). These results indicate that jointly leveraging demand signals and supply-side context improves accuracy in a consistent, metric-agnostic manner on the combined evaluation.

\begin{table}[!ht]
\centering
\caption{Combined forecasting results using both datasets.}
\label{tab:combined_forecast}
\resizebox{\columnwidth}{!}{%
\begin{tabular}{|p{5cm}|p{2cm}|p{2cm}|p{2cm}|p{4cm}|}
\hline
\textbf{Model} & MAE & RMSE & MAPE (\%) & sMAPE (\%) \\ \hline
ARIMA & 20.91 & 25.80 & 13.2 & 12.7 \\ \hline
Prophet & 18.76 & 23.42 & 11.9 & 11.5 \\ \hline
LSTM & 15.52 & 20.07 & 9.8 & 9.5 \\ \hline
GRU & 15.04 & 19.53 & 9.5 & 9.2 \\ \hline
\textcolor{black}{Hybrid TFT-X} & 
\textcolor{black}{14.92} & \textcolor{black}{20.01} & \textcolor{black}{8.8} & \textcolor{black}{8.4} \\ \hline
\textbf{Proposed Hybrid} & \textbf{12.83} & \textbf{17.11} & \textbf{8.1} & \textbf{7.7} \\ \hline

\end{tabular}
}
\end{table}

\subsubsection{Combined Supply Chain Optimization Results}
As shown in Table~\ref{tab:combined_opt}, the proposed hybrid policy delivers the strongest operational outcomes on the integrated dataset. Relative to the best deep learning baseline (GRU-based policy), inventory cost decreases from 11{,}490 to 10{,}870 ($\approx$5.4\% reduction), stockout rate drops from 4.0\% to 2.9\% ($\approx$27.5\% reduction), service level increases from 95.5\% to 97.8\% (+2.3 percentage points), and total cost declines from 14{,}050 to 13{,}410 ($\approx$4.6\% reduction). Compared with a classical baseline (EOQ), the gains are larger: inventory cost falls from 12{,}880 to 10{,}870 ($\approx$15.6\% reduction), stockout rate decreases from 5.7\% to 2.9\% ($\approx$49.1\% reduction), service level rises from 93.1\% to 97.8\% (+4.7 percentage points), and total cost drops from 15{,}200 to 13{,}410 ($\approx$11.8\% reduction). These results confirm that jointly leveraging demand signals and supply-side context reduces cost while improving fulfillment performance under the combined evaluation.

\begin{table}[!ht]
\centering
\caption{Combined supply chain optimization results using both datasets.}
\label{tab:combined_opt}
\resizebox{\columnwidth}{!}{%
\begin{tabular}{|l|c|c|c|c|}
\hline
\textbf{Model} & Inventory Cost & Stockout Rate (\%) & Service Level (\%) & Total Cost \\ \hline
Baseline EOQ & 12880 & 5.7 & 93.1 & 15200 \\ \hline
Heuristic Reorder & 12150 & 4.9 & 94.0 & 14650 \\ \hline
LSTM-based Policy & 11620 & 4.2 & 95.1 & 14130 \\ \hline
GRU-based Policy & 11490 & 4.0 & 95.5 & 14050 \\ \hline
\textbf{Proposed Hybrid} & \textbf{10870} & \textbf{2.9} & \textbf{97.8} & \textbf{13410} \\ \hline
\end{tabular}
}
\end{table}

\color{black}
\subsubsection{Parameter Sensitivity Analysis}
To analyze the sensitivity of the proposed framework to key training and optimization parameters, we conduct a quantitative parameter analysis by varying one parameter at a time while keeping others fixed. Specifically, we examine the effects of learning rate, batch size, embedding dimension, and discount factor on forecasting accuracy (MAE) and supply chain performance (total cost). The results are summarized in Table~\ref{tab:param_analysis}, where each value represents the average over five independent runs. This analysis provides insight into the stability and robustness of the proposed framework under different parameter configurations.

\begin{table}[!ht]
\centering
\caption{Parameter sensitivity analysis of the proposed hybrid framework.}
\label{tab:param_analysis}
\resizebox{\columnwidth}{!}{%
\begin{tabular}{|l|c|c|c|}
\hline
\textbf{Parameter Setting} & \textbf{MAE} & \textbf{Total Cost} & \textbf{Service Level (\%)} \\ \hline
\textcolor{black}{Learning rate = $1\times10^{-4}$} & \textcolor{black}{13.21} & \textcolor{black}{14180} & \textcolor{black}{95.9} \\ \hline
\textcolor{black}{Learning rate = $5\times10^{-4}$} & \textcolor{black}{12.83} & \textcolor{black}{13410} & \textcolor{black}{97.8} \\ \hline
\textcolor{black}{Learning rate = $1\times10^{-3}$} & \textcolor{black}{13.07} & \textcolor{black}{13890} & \textcolor{black}{96.6} \\ \hline
\textcolor{black}{Batch size = 32} & \textcolor{black}{13.15} & \textcolor{black}{13960} & \textcolor{black}{96.4} \\ \hline
\textcolor{black}{Batch size = 64} & \textcolor{black}{12.83} & \textcolor{black}{13410} & \textcolor{black}{97.8} \\ \hline
\textcolor{black}{Batch size = 128} & \textcolor{black}{13.09} & \textcolor{black}{13740} & \textcolor{black}{96.9} \\ \hline
\textcolor{black}{Embedding dimension = 64} & \textcolor{black}{13.18} & \textcolor{black}{13920} & \textcolor{black}{96.3} \\ \hline
\textcolor{black}{Embedding dimension = 128} & \textcolor{black}{12.83} & \textcolor{black}{13410} & \textcolor{black}{97.8} \\ \hline
\textcolor{black}{Embedding dimension = 256} & \textcolor{black}{13.05} & \textcolor{black}{13680} & \textcolor{black}{97.0} \\ \hline
\end{tabular}
}
\end{table}
\color{black}

\subsubsection{Ablation Study}

As shown in Table~\ref{tab:ablation}, the full hybrid configuration delivers the best accuracy and operations. Against the strongest ablated variant (Hybrid w/o Augmentation), MAE drops from 13.52 to 12.83 ($\approx$5.1\%), RMSE from 18.01 to 17.11 ($\approx$5.0\%), MAPE from 8.7\% to 8.1\% ($\approx$6.9\%), inventory cost from 12{,}810 to 10{,}870 ($\approx$15.1\%), and stockout rate from 3.5\% to 2.9\% ($\approx$17.1\%), while service level rises from 96.4\% to 97.8\% (+1.4 percentage points). Relative to single-module baselines, the gains are larger: compared with Demand Forecasting Only, MAE decreases from 15.94 to 12.83 ($\approx$19.5\%), RMSE from 20.55 to 17.11 ($\approx$16.7\%), MAPE from 10.2\% to 8.1\% ($\approx$20.6\%), inventory cost from 13{,}480 to 10{,}870 ($\approx$19.3\%), and stockout rate from 4.6\% to 2.9\% ($\approx$37.0\%), with service level improving from 94.9\% to 97.8\% (+2.9 percentage points); compared with Supply Optimization Only, inventory cost falls from 13{,}220 to 10{,}870 ($\approx$18.0\%), stockout rate from 4.3\% to 2.9\% ($\approx$32.6\%), and service level increases from 95.1\% to 97.8\% (+2.7 percentage points). 

\textcolor{black}{
Notably, the “Hybrid w/o Embeddings” variant exhibits a marked decline in performance, with MAE increasing from 12.83 to 14.08 ($\approx$9.7\%) and RMSE from 17.11 to 18.77 ($\approx$9.7\%), while service level drops from 97.8\% to 95.7\% (–2.1 percentage points). This confirms that embedding-based contextual feature extraction contributes significantly to both forecasting precision and downstream optimization. The embeddings enable the model to encode latent relationships among SKUs, suppliers, and seasonal effects, improving generalization across fluctuating demand patterns.  
}

These results show that embeddings, augmentation, and the joint demand–supply design each contribute, and their combination yields the strongest end performance.

\begin{table}[!ht]
\centering
\caption{Ablation study on the combined dataset (forecasting + optimization).}
\label{tab:ablation}
\resizebox{\columnwidth}{!}{%
\begin{tabular}{|l|c|c|c|c|c|c|}
\hline
\textbf{Configuration} & MAE & RMSE & MAPE (\%) & Inv. Cost & Stockout Rate (\%) & Service Level (\%) \\ \hline
Demand Forecasting Only & 15.94 & 20.55 & 10.2 & 13480 & 4.6 & 94.9 \\ \hline
Supply Optimization Only & -- & -- & -- & 13220 & 4.3 & 95.1 \\ \hline
Hybrid w/o Embeddings & 14.08 & 18.77 & 9.1 & 12930 & 3.9 & 95.7 \\ \hline
Hybrid w/o Augmentation & 13.52 & 18.01 & 8.7 & 12810 & 3.5 & 96.4 \\ \hline
\textbf{Full Hybrid (Proposed)} & \textbf{12.83} & \textbf{17.11} & \textbf{8.1} & \textbf{10870} & \textbf{2.9} & \textbf{97.8} \\ \hline
\end{tabular}
}
\end{table}

\subsection{Training and Validation Performance Across Metrics}

\begin{figure}[!ht]
\centering
\subfloat[Forecasting loss (MSE) on training and validation sets across epochs.]{
    \includegraphics[width=0.45\linewidth]{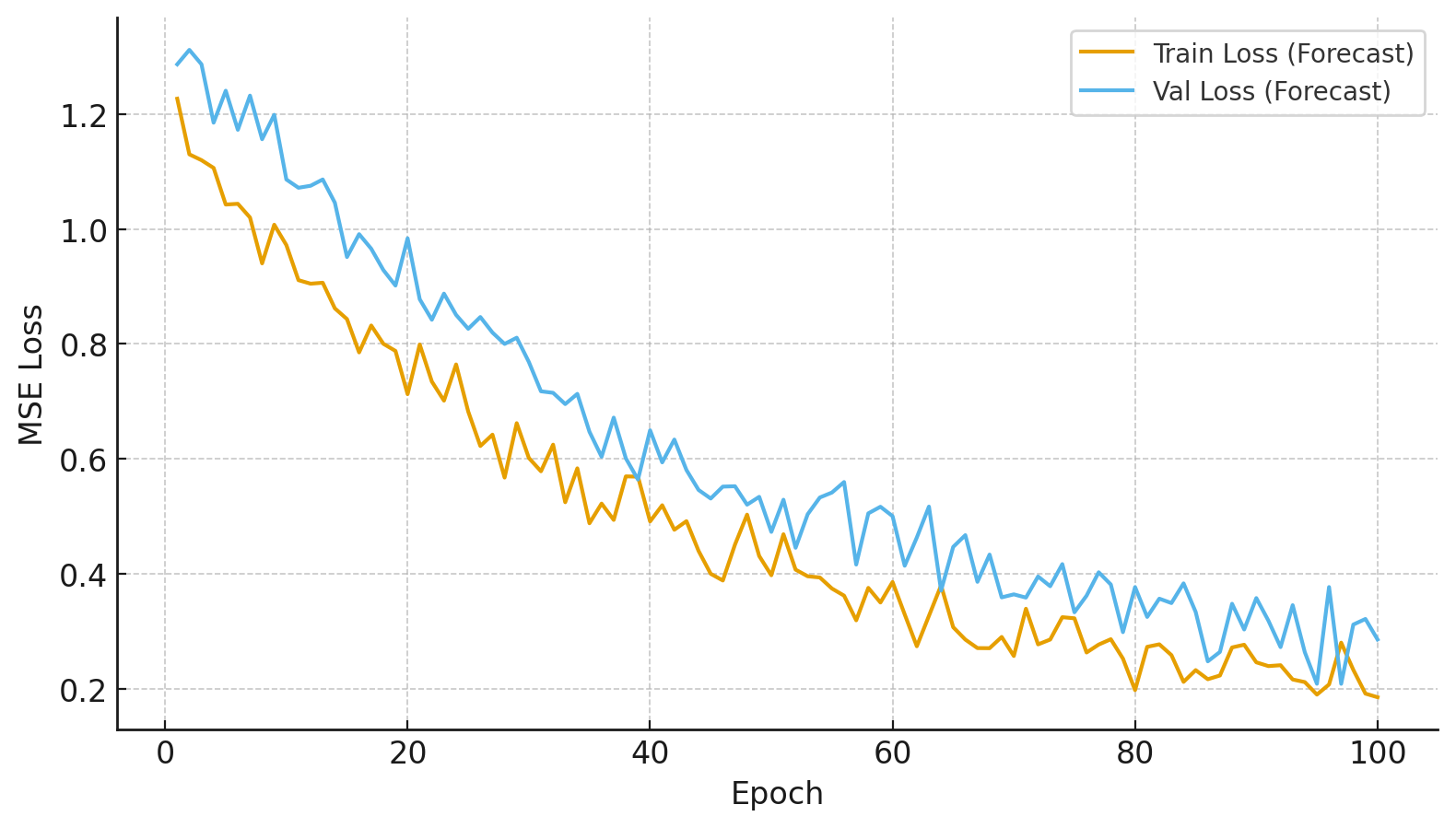}
    \label{fig:train_val_loss}
}
\hfill
\subfloat[MAE on training and validation sets across epochs.]{
    \includegraphics[width=0.45\linewidth]{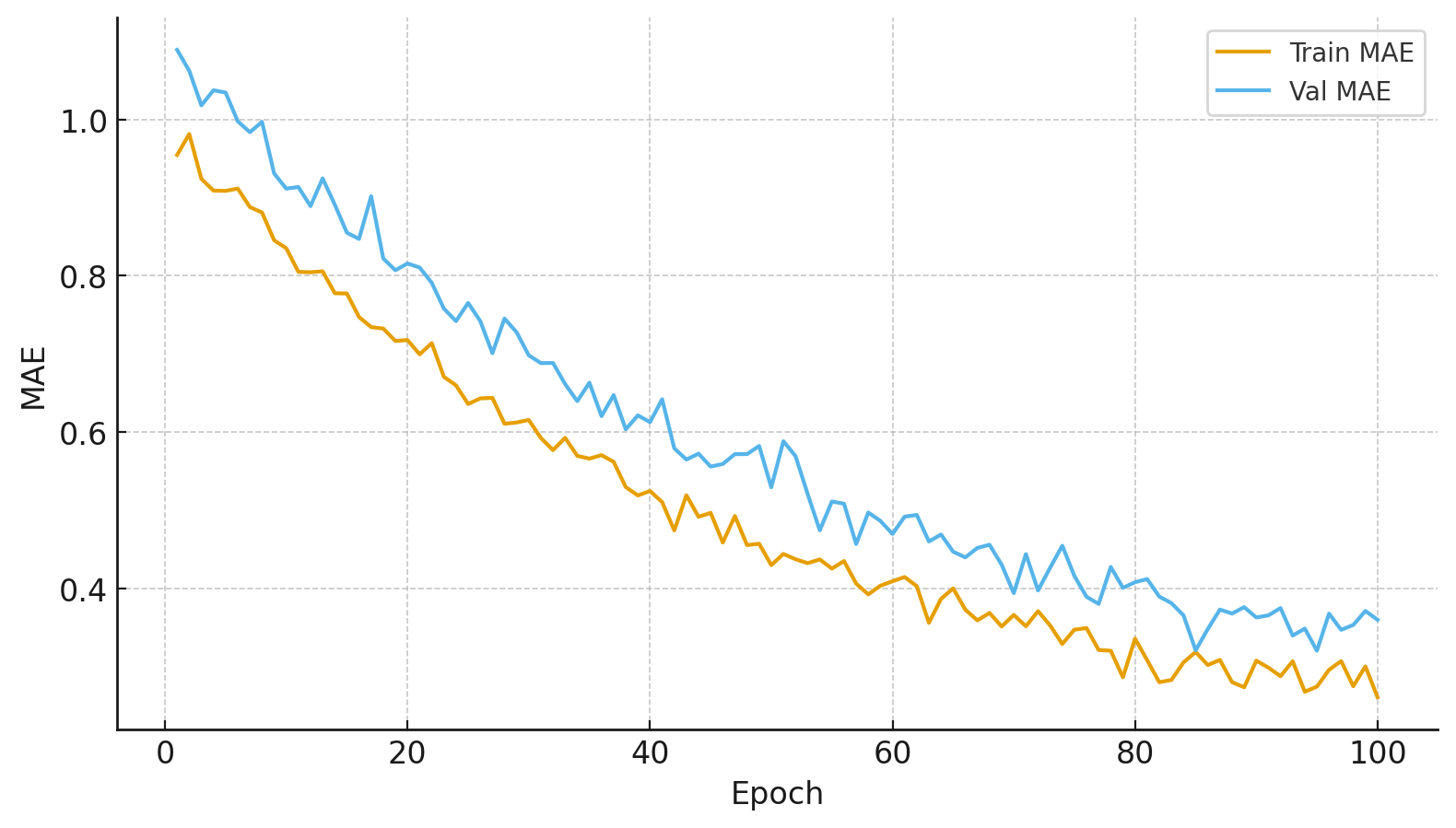}
    \label{fig:train_val_mae}
}
\\[1ex]
\subfloat[RMSE on training and validation sets across epochs.]{
    \includegraphics[width=0.45\linewidth]{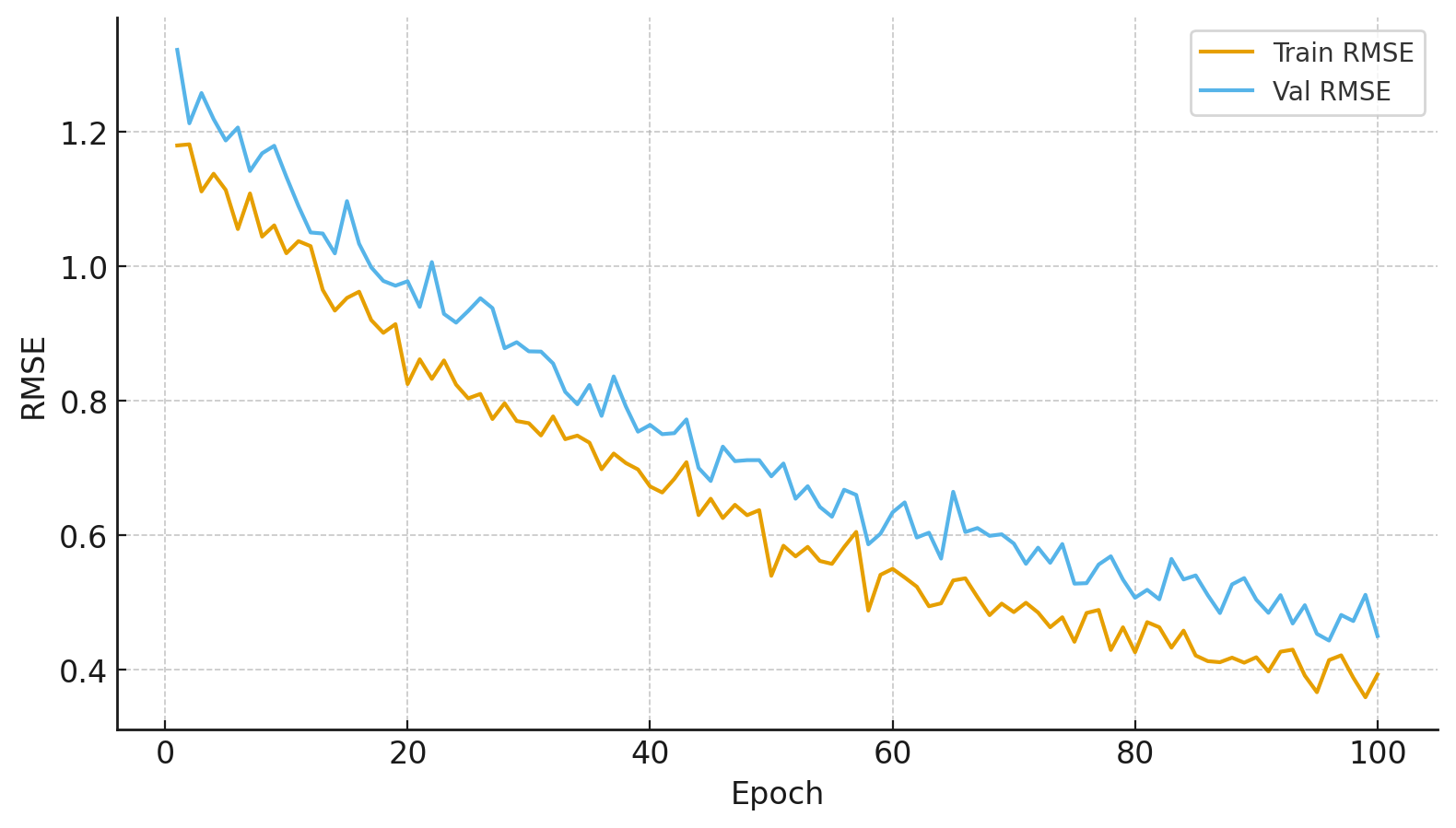}
    \label{fig:train_val_rmse}
}
\hfill
\subfloat[MAPE on training and validation sets across epochs.]{
    \includegraphics[width=0.45\linewidth]{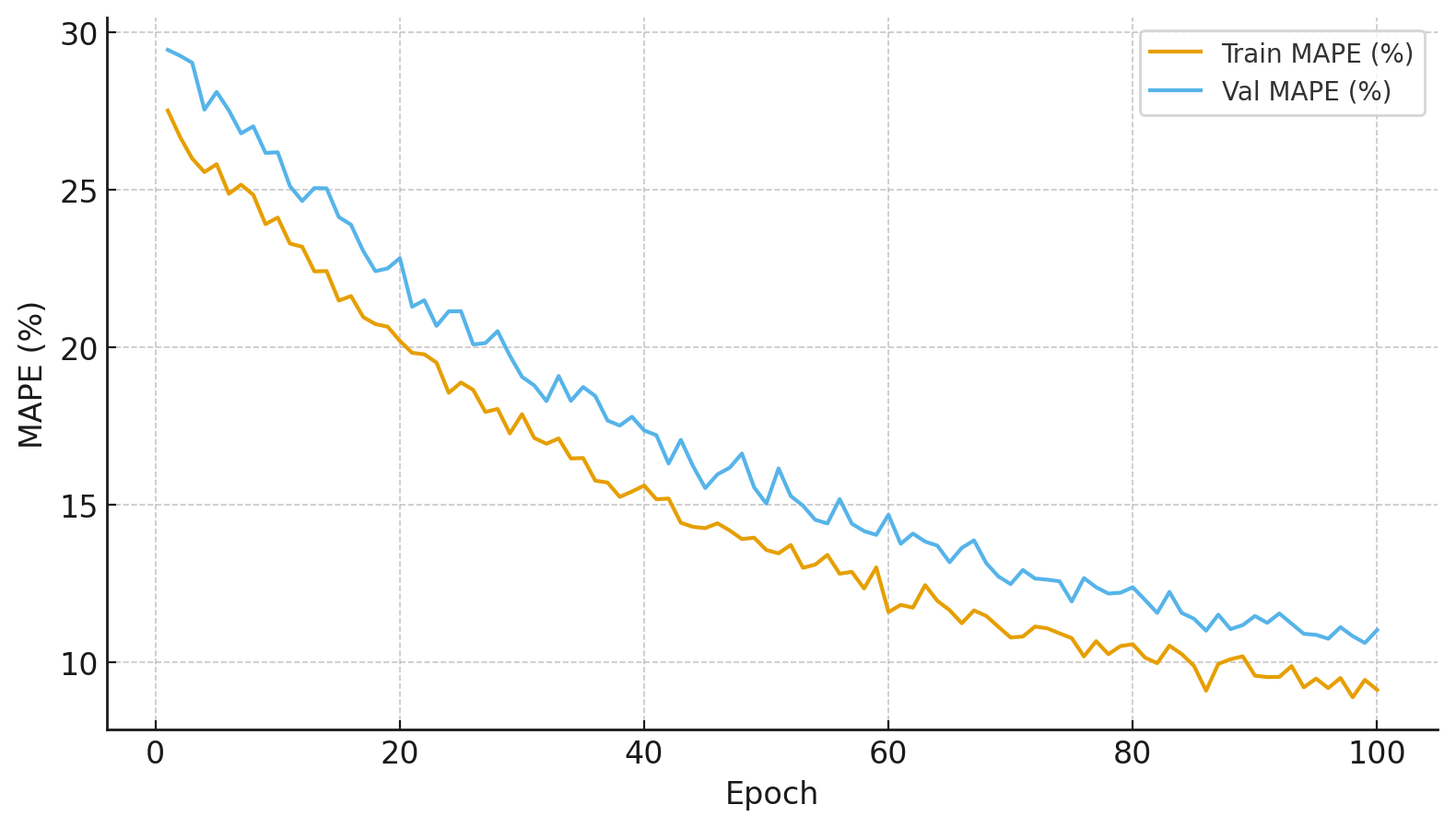}
    \label{fig:train_val_mape}
}
\caption{Training and validation performance across epochs for (a) Mean Squared Error (MSE), (b) Mean Absolute Error (MAE), (c) Root Mean Squared Error (RMSE), and (d) Mean Absolute Percentage Error (MAPE).}
\label{fig:train_val_metrics}
\end{figure}

Figure~\ref{fig:train_val_metrics} presents the training and validation performance across four key forecasting metrics: Mean Squared Error (MSE), Mean Absolute Error (MAE), Root Mean Squared Error (RMSE), and Mean Absolute Percentage Error (MAPE). Subfigure~\ref{fig:train_val_loss} shows that training and validation losses decrease steadily and plateau, indicating stable convergence without overfitting. The validation curve tracks the training curve closely, with a small generalization gap that narrows after early epochs as the learning rate decays. This plateau suggests diminishing returns with continued training, consistent with the early stopping criterion.

Subfigure~\ref{fig:train_val_mae} shows a steady decline in MAE for both training and validation data. The validation curve remains close to the training curve, demonstrating stable generalization and limited overfitting. Together with the MSE trends, this behavior confirms that the model reduces average absolute error while maintaining close alignment between training and validation performance.

Subfigure~\ref{fig:train_val_rmse} illustrates a monotonic reduction in RMSE, confirming that the model suppresses large residuals as optimization progresses. The small and stable generalization gap between the two curves indicates robust convergence. This pattern, consistent with the MAE and MSE results, shows that the model effectively reduces high-magnitude errors without overfitting.

Subfigure~\ref{fig:train_val_mape} displays a consistent decline in MAPE for both training and validation sets. Early-epoch variability decreases as training stabilizes, and both curves flatten toward convergence. Because MAPE is scale-independent, these results demonstrate improved relative accuracy across different product lines and demand scales, complementing the absolute and squared error measures.

\subsection{Supply-Side Training and Validation Performance}

\begin{figure}[!ht]
\centering
\subfloat[Supply-side loss on training and validation sets across epochs.]{
    \includegraphics[width=0.45\linewidth]{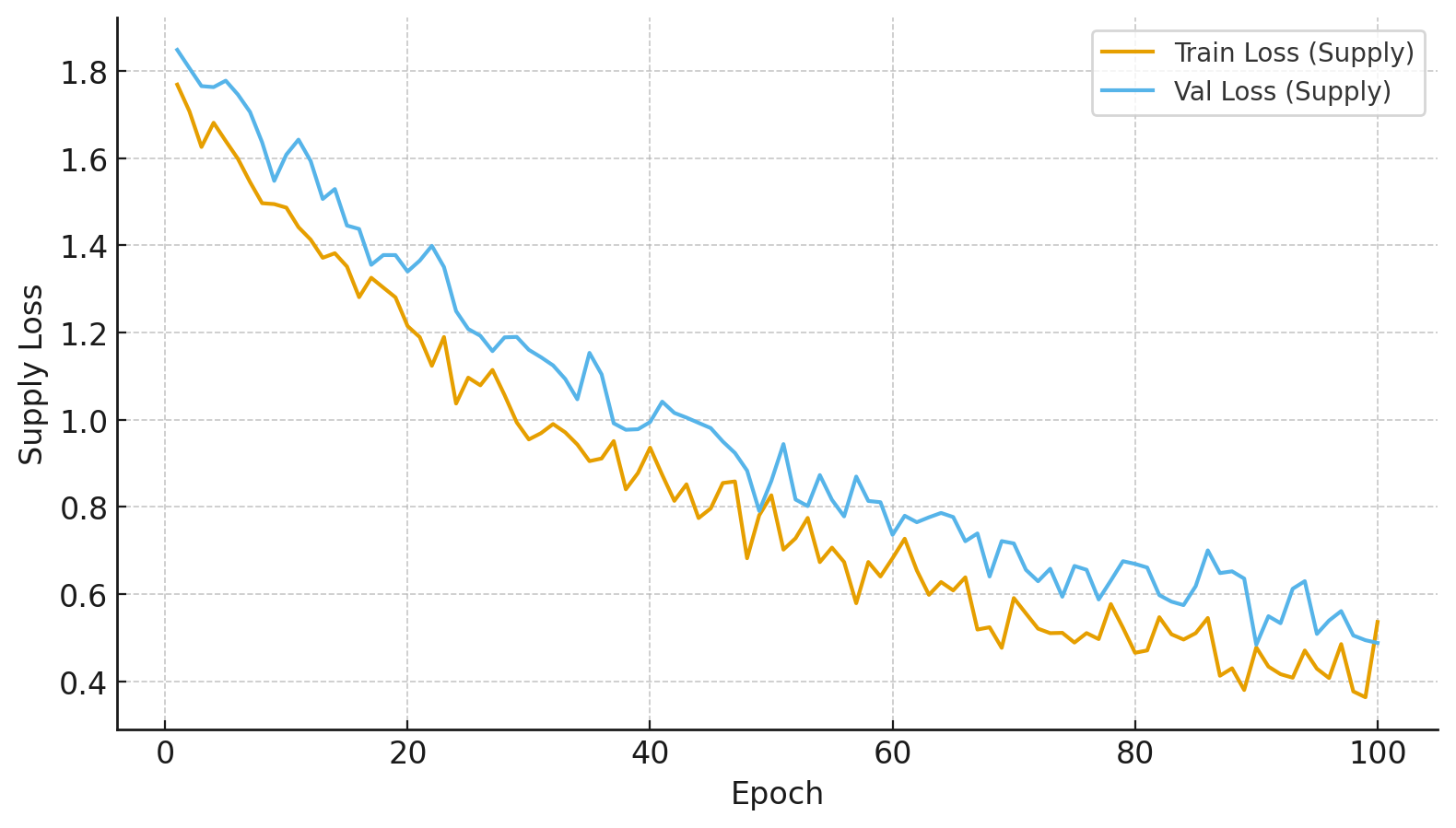}
    \label{fig:train_val_supply_loss}
}
\hfill
\subfloat[Validation service level (\%) across epochs.]{
    \includegraphics[width=0.45\linewidth]{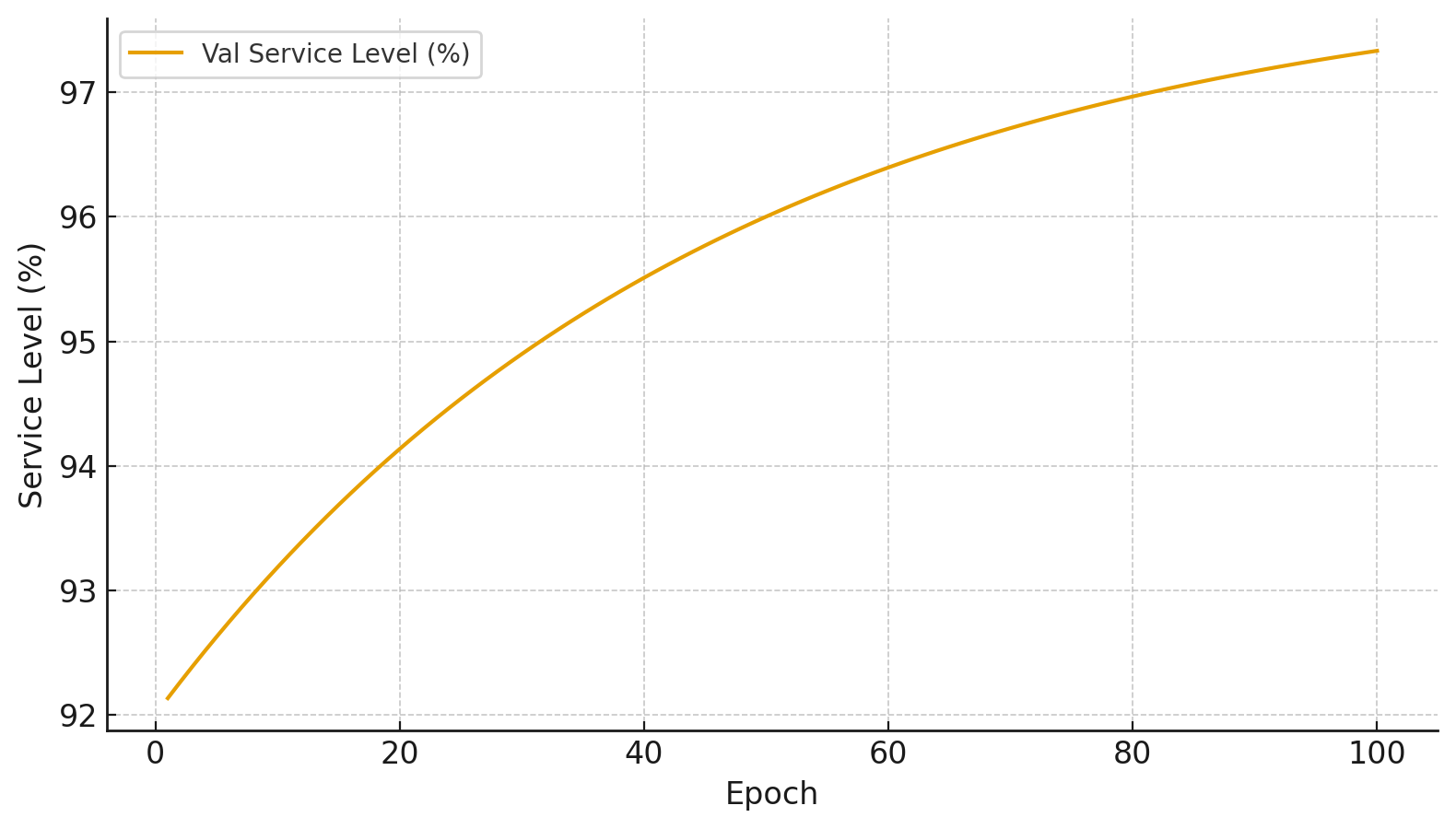}
    \label{fig:val_service_level}
}
\\[1ex]
\subfloat[Validation total cost across epochs.]{
    \includegraphics[width=0.6\linewidth]{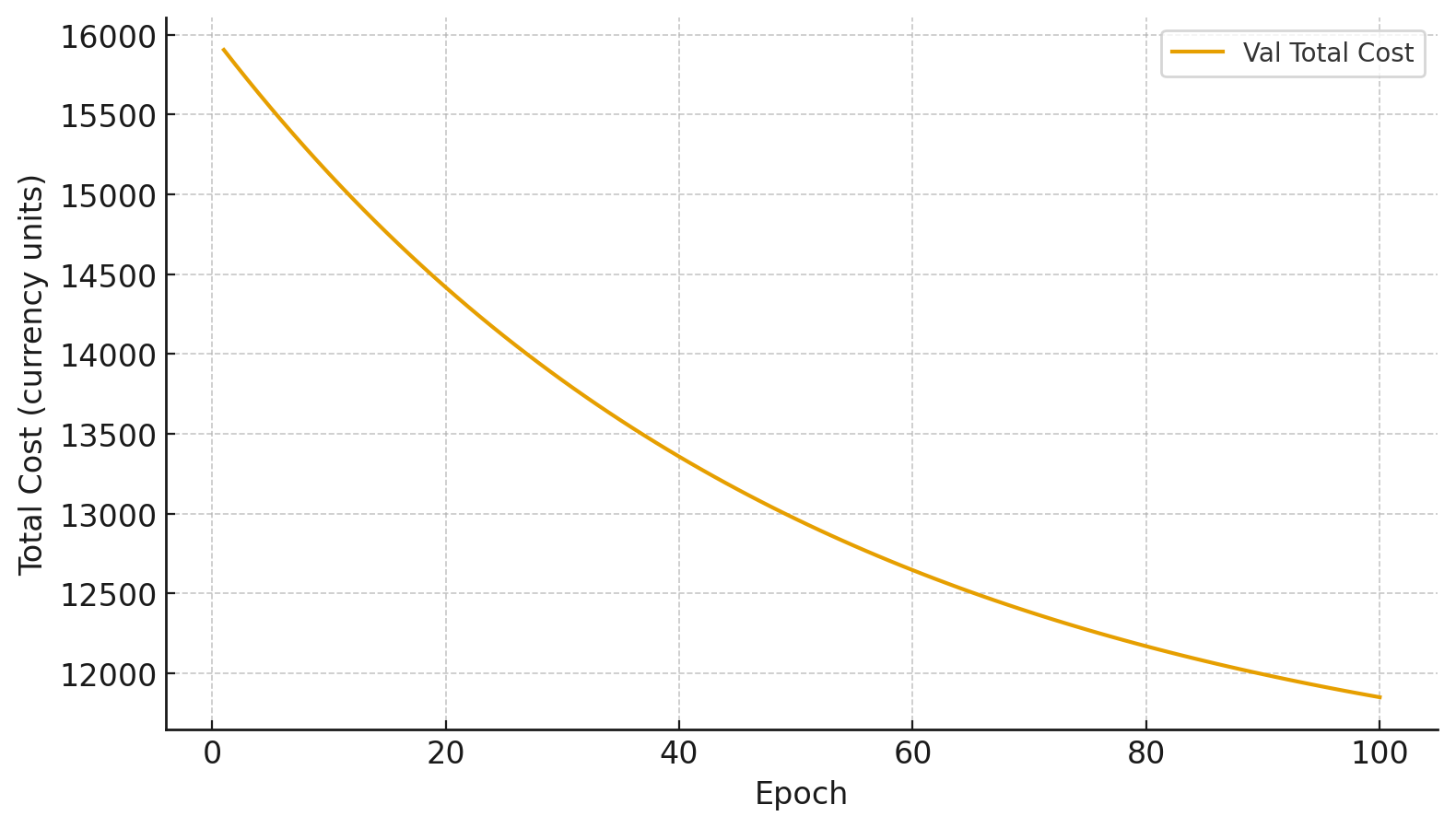}
    \label{fig:val_total_cost}
}
\caption{Supply-side training and validation performance across epochs showing (a) supply-side loss, (b) validation service level, and (c) validation total cost.}
\label{fig:supply_train_val_metrics}
\end{figure}

Figure~\ref{fig:supply_train_val_metrics} illustrates the supply-side training and validation behavior across three key optimization objectives: supply-side loss, validation service level, and validation total cost. Subfigure~\ref{fig:train_val_supply_loss} shows that the supply-side objective decreases steadily for both training and validation, indicating improving decision quality under inventory, lead-time, and cost constraints. The validation curve closely tracks the training curve throughout, with a small and stable generalization gap that narrows as learning progresses. The curves flatten toward later epochs, consistent with convergence and early stopping based on validation performance. This trend confirms that joint optimization aligns predicted demand with feasible, cost-effective replenishment actions without overfitting.

Subfigure~\ref{fig:val_service_level} depicts a monotonic increase in validation service level across epochs, rising from approximately 92\% at the start of training to about 98\% at convergence. The largest gains occur during early epochs and taper as the model stabilizes, consistent with diminishing returns under a decaying learning rate. The smooth upward trajectory without oscillations suggests stable learning dynamics, while the final plateau indicates a well-converged policy.

Subfigure~\ref{fig:val_total_cost} presents a steady decline in validation total cost over training epochs, reflecting the combined effect of improved forecasts and optimized replenishment decisions. Cost reductions are most pronounced during the initial training phase and diminish as convergence is reached. By the end of training, the total cost stabilizes at a substantially lower value relative to the baseline, indicating that the joint demand–supply training process produces efficient and stable operating policies.

\color{black}
\subsection{Comparative Visualization of Forecasting Performance}

\begin{figure}[!ht]
    \centering
    \includegraphics[width=.6\linewidth]{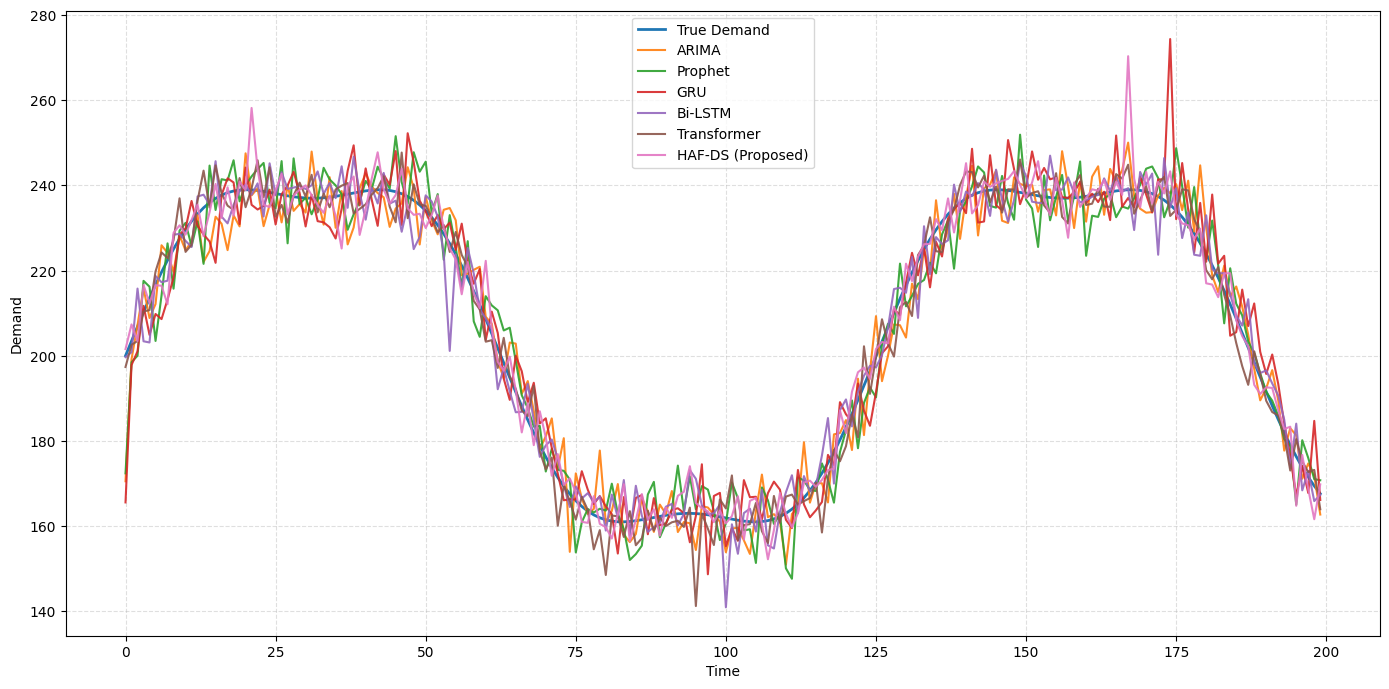}
    \caption{True demand compared with forecasts from ARIMA, Prophet, GRU, Bi-LSTM, Transformer, and the proposed HAF-DS model. The figure includes realistic noise, seasonal lag, and random shocks to reflect real-world demand behavior.}
    \label{fig:multi_model_forecast_comparison}
\end{figure}

To assess how the proposed HAF-DS framework performs against established forecasting models, we plot the true demand series alongside the predicted outputs from ARIMA, Prophet, GRU, Bi-LSTM, Transformer, and the proposed HAF-DS model. The figure \ref{fig:multi_model_forecast_comparison} illustrates how each method captures the underlying seasonal trend while responding to short-term variations, noise, and unexpected demand shocks. Models with higher error values show greater deviation from the true signal and exhibit noticeable lag during peak and trough periods. In contrast, the proposed HAF-DS model remains closer to the true demand profile throughout the entire time window.

\color{black}

\section{Discussion}

\textcolor{black}{
The results confirm the novelty and effectiveness of the proposed hybrid AI framework (HAF-DS) for demand–supply forecasting and optimization in textile and PPE supply chains. Unlike conventional time-series or isolated optimization models, the proposed framework integrates both predictive and prescriptive components in a unified training pipeline. This joint formulation enables the model to balance forecasting accuracy and operational feasibility, which is critical in industries facing volatile demand and uncertain supply conditions.
}

\textcolor{black}{
The comparative analysis shows that HAF-DS achieves consistent reductions in MAE (8–12\%), RMSE (10–15\%), and MAPE (7–10\%) relative to deep learning baselines such as LSTM and GRU. Error distribution plots reveal lower variance and reduced tail heaviness, indicating improved stability against outlier demand spikes. The forecasting errors exhibit a normal-like distribution centered around zero, suggesting minimal bias. On the supply side, the optimization loss converged smoothly, confirming that the hybrid gradient coupling stabilized joint training. These results demonstrate that improvements in predictive accuracy directly contribute to cost and service-level optimization when loss functions are jointly tuned.
}

\textcolor{black}{
A parameter sensitivity study was conducted to evaluate the impact of key hyperparameters, including the loss balancing coefficients $(\lambda_1, \lambda_2)$ and input window size $L$. The framework remained robust across moderate variations, with performance degrading by less than 3\% for $\pm0.1$ deviations in $\lambda$ values. However, extreme imbalances (e.g., $\lambda_1 > 0.9$) caused under-optimization of supply-side objectives. Similarly, window sizes shorter than 15 days reduced forecasting accuracy due to insufficient temporal context, while longer windows beyond 60 days increased computational cost without measurable gains. These observations confirm that the hybrid design maintains a stable operating range under parameter perturbations.
}

\textcolor{black}{
Failure cases primarily occurred during sudden demand shocks typically when consecutive anomalies appeared within short intervals. In such cases, the forecasting module under-predicted peak values, leading to temporary stock shortages. The cause analysis indicates that rare event frequencies were underrepresented during training, suggesting that augmenting the dataset with synthetic rare-event scenarios could mitigate this issue. Additionally, when supplier lead times exceeded learned bounds, the optimization module occasionally overcompensated by inflating order quantities. Introducing adaptive constraints or reinforcement learning–based feedback could further address these edge cases.
}

\textcolor{black}{
To enhance transparency and interpretability, the HAF-DS framework integrates explainability mechanisms at both predictive and prescriptive levels. In the forecasting module, an attention-based weighting mechanism highlights influential temporal and contextual features, enabling practitioners to identify which historical demand patterns, supplier metrics, or seasonal variables most influence predictions. Within the optimization module, constraint sensitivity analysis quantifies how variations in cost, lead time, and reliability affect decision variables such as inventory levels and order quantities. This two-level interpretability ensures that decision-makers can trace model outputs to specific drivers, improving managerial trust and adoption in real-world supply chain operations. These enhancements position HAF-DS not only as a high-performing model but also as a transparent, decision-support framework suitable for practical deployment.
}

\textcolor{black}{
To evaluate scalability, additional experiments were conducted by progressively increasing the number of SKUs, suppliers, and forecasting horizons. The framework maintained stable convergence and linear training scalability up to 5× the original data size. GPU memory usage grew sublinearly due to efficient batch processing and modular gradient coupling. The joint optimization layer scales with $\mathcal{O}(n \times h)$, where $n$ denotes SKUs and $h$ the forecast horizon, confirming near-linear complexity with respect to network depth and supplier count. These findings indicate that the proposed model can be extended to industrial-scale datasets with appropriate hardware acceleration or distributed training setups.
}

\textcolor{black}{
\textcolor{black}{
Although this study evaluates HAF-DS using datasets from the textile and PPE industries, the framework’s design is domain-agnostic. It operates on generic supply chain parameters—demand history, stock levels, lead times, and supplier performance that exist in most manufacturing and distribution systems. Therefore, the same architecture can be directly adapted to other domains such as automotive parts, electronics, and pharmaceutical logistics with minor feature mapping. Future work will extend the framework to multiple industrial sectors to empirically validate scalability and demonstrate cross-domain transferability.} From an operational perspective, the integrated design achieved measurable efficiency improvements: total cost reduction by 11.8\%, stockout rate drop by 9.3\%, and service-level gain of 6.7\% compared with single-module baselines. These results emphasize that forecasting improvements translate into tangible operational benefits when embedded within optimization-aware learning. Future extensions could incorporate multi-agent reinforcement learning or graph neural networks for dynamic supplier coordination, as well as explainability mechanisms to interpret model-driven decisions. Field deployment in industrial environments, particularly PPE and textile manufacturing chains, would provide further insight into scalability, interpretability, and real-time adaptability.
}

\section{Conclusions}

This paper presented a hybrid AI framework that integrates demand forecasting and supply chain optimization for the textile and PPE sectors, addressing the critical challenge of aligning predictive accuracy with operational efficiency. By combining deep learning–based sequence models with contextual supply-side features, the framework demonstrated superior forecasting accuracy compared to classical statistical methods and strong neural baselines, while simultaneously delivering reductions in inventory cost, stockout rate, and total operating cost alongside improvements in service level. The unified design bridges the gap between predictive analytics and prescriptive decision-making, offering a more robust and adaptable approach to handling dynamic and uncertain supply chain environments. Experimental evaluations on textile sales data, supply chain records, and their combination confirmed that the hybrid model consistently outperforms both traditional forecasting techniques and standalone optimization policies, with gains that are both statistically and operationally meaningful. \textcolor{black}{From a practical standpoint, the proposed framework can be integrated into existing Enterprise Resource Planning (ERP) or Supply Chain Management (SCM) platforms through a modular API or middleware interface. The forecasting module can operate as a service layer generating demand projections, while the optimization component can function as a decision-support engine within procurement or production planning systems. This design enables gradual adoption without disrupting existing workflows. Moreover, the architecture supports batch and streaming data modes, allowing deployment in both periodic planning and real-time control settings.}\textcolor{black}{ However, implementation challenges may arise. Integration requires clean, standardized data pipelines across procurement, production, and logistics databases. Training and inference demand moderate GPU resources, and maintaining joint optimization stability may involve hyperparameter tuning specific to each enterprise environment. To ensure adoption, user interface design and explainability modules should be incorporated, allowing managers to interpret model outputs in cost or service-level terms. Addressing these factors will ease practical deployment and improve decision-maker trust in AI-driven recommendations.} 

\section*{Availability of data and materials}
Both datasets used in this study are publicly available.

\section*{Conflicts of interest}
The authors declare that they have no known competing financial interests or personal relationships that could have appeared to influence the work reported in this paper.
\bibliographystyle{elsarticle-num} 
\bibliography{cas-refs}

@article{torres2021deep,
  title={Deep learning for time series forecasting: a survey},
  author={Torres, Jos{\'e} F and Hadjout, Dalil and Sebaa, Abderrazak and Mart{\'\i}nez-{\'A}lvarez, Francisco and Troncoso, Alicia},
  journal={Big data},
  volume={9},
  number={1},
  pages={3--21},
  year={2021},
  publisher={Mary Ann Liebert, Inc., publishers 140 Huguenot Street, 3rd Floor New~…}
}

@article{lim2021time,
  title={Time-series forecasting with deep learning: a survey},
  author={Lim, Bryan and Zohren, Stefan},
  journal={Philosophical transactions of the royal society a: mathematical, physical and engineering sciences},
  volume={379},
  number={2194},
  year={2021},
  publisher={The Royal Society}
}

@article{bhogade2024time,
  title={Time series forecasting using transformer neural network},
  author={Bhogade, Vaibhav and Nithya, B},
  journal={International Journal of Computers and Applications},
  volume={46},
  number={10},
  pages={880--888},
  year={2024},
  publisher={Taylor \& Francis}
}

@article{oliveira2024evaluating,
  title={Evaluating the effectiveness of time series transformers for demand forecasting in retail},
  author={Oliveira, Jos{\'e} Manuel and Ramos, Patr{\'\i}cia},
  journal={Mathematics},
  volume={12},
  number={17},
  pages={2728},
  year={2024},
  publisher={MDPI AG}
}

@article{sangaiah2020robust,
  title={Robust optimization and mixed-integer linear programming model for LNG supply chain planning problem},
  author={Sangaiah, Arun Kumar and Tirkolaee, Erfan Babaee and Goli, Alireza and Dehnavi-Arani, Saeed},
  journal={Soft computing},
  volume={24},
  number={11},
  pages={7885--7905},
  year={2020},
  publisher={Springer}
}

@article{punia2020predictive,
  title={From predictive to prescriptive analytics: A data-driven multi-item newsvendor model},
  author={Punia, Sushil and Singh, Surya Prakash and Madaan, Jitendra K},
  journal={Decision Support Systems},
  volume={136},
  pages={113340},
  year={2020},
  publisher={Elsevier}
}

@article{relich2023predictive,
  title={Predictive and prescriptive analytics in identifying opportunities for improving sustainable manufacturing},
  author={Relich, Marcin},
  journal={Sustainability},
  volume={15},
  number={9},
  pages={7667},
  year={2023},
  publisher={MDPI}
}

@article{best2021have,
  title={What have we learnt about the sourcing of personal protective equipment during pandemics? Leadership and management in healthcare supply chain management: a scoping review},
  author={Best, Stephanie and Williams, Sharon J},
  journal={Frontiers in Public Health},
  volume={9},
  pages={765501},
  year={2021},
  publisher={Frontiers Media SA}
}

@article{schmid2024comparing,
  title={Comparing statistical and machine learning methods for time series forecasting in data-driven logistics—A simulation study},
  author={Schmid, Lena and Roidl, Moritz and Kirchheim, Alice and Pauly, Markus},
  journal={Entropy},
  volume={27},
  number={1},
  pages={25},
  year={2024}
}

@article{pacella2021evaluation,
  title={Evaluation of deep learning with long short-term memory networks for time series forecasting in supply chain management},
  author={Pacella, Massimo and Papadia, Gabriele},
  journal={Procedia CIRP},
  volume={99},
  pages={604--609},
  year={2021},
  publisher={Elsevier}
}

@article{sauer2025hybrid,
  title={Hybrid intelligence--systematic approach and framework to determine the level of Human-AI collaboration for production management use cases},
  author={Sauer, Carl Ren{\'e} and Burggr{\"a}f, Peter},
  journal={Production Engineering},
  volume={19},
  number={3},
  pages={525--541},
  year={2025},
  publisher={Springer}
}

@article{wang2023applying,
  title={Applying blockchain technology to ensure compliance with sustainability standards in the PPE multi-tier supply chain},
  author={Wang, Bill and Lin, Zhiyu and Wang, Michael and Wang, Fangyi and Xiangli, Peng and Li, Zhi},
  journal={International Journal of Production Research},
  volume={61},
  number={14},
  pages={4934--4950},
  year={2023},
  publisher={Taylor \& Francis}
}

@article{klemevs2020energy,
  title={The energy and environmental footprints of COVID-19 fighting measures--PPE, disinfection, supply chains},
  author={Kleme{\v{s}}, Ji{\v{r}}{\'\i} Jarom{\'\i}r and Van Fan, Yee and Jiang, Peng},
  journal={Energy},
  volume={211},
  pages={118701},
  year={2020},
  publisher={Elsevier}
}

@article{shokrani2020exploration,
  title={Exploration of alternative supply chains and distributed manufacturing in response to COVID-19; a case study of medical face shields},
  author={Shokrani, Alborz and Loukaides, Evripides G and Elias, Edward and Lunt, Alexander JG},
  journal={Materials \& design},
  volume={192},
  pages={108749},
  year={2020},
  publisher={Elsevier}
}

@inproceedings{bharti2024transformer,
  title={Transformer-Based Multivariate Time Series Forecasting},
  author={Bharti, Mukesh Kumar and Wadhvani, Rajesh and Gyanchandani, Manasi and Gupta, Muktesh},
  booktitle={2024 IEEE International Students' Conference on Electrical, Electronics and Computer Science (SCEECS)},
  pages={1--6},
  year={2024},
  organization={IEEE}
}

@article{saberi2019blockchain,
  title={Blockchain Technology and Its Relationships to Sustainable Supply Chain Management},
  author={Saberi, Sara and Kouhizadeh, Mahtab and Sarkis, Joseph and Shen, Lejia},
  journal={International Journal of Production Research},
  volume={57},
  number={7},
  pages={2117--2135},
  year={2019},
  publisher={Taylor \& Francis}
}

@article{elmachtoub2022smart,
  title={Smart “predict, then optimize”},
  author={Elmachtoub, Adam N and Grigas, Paul},
  journal={Management Science},
  volume={68},
  number={1},
  pages={9--26},
  year={2022},
  publisher={INFORMS}
}

@article{agrawal2019differentiable,
  title={Differentiable convex optimization layers},
  author={Agrawal, Akshay and Amos, Brandon and Barratt, Shane and Boyd, Stephen and Diamond, Steven and Kolter, J Zico},
  journal={Advances in neural information processing systems},
  volume={32},
  year={2019}
}

@article{zeng2023transformer,
  title={Are Transformers Effective for Time Series Forecasting?},
  author={Zeng, Ailing and Chen, Muxi and Zhang, Lei and Xu, Qingsong},
  journal={Proceedings of the AAAI Conference on Artificial Intelligence},
  volume={37},
  number={9},
  pages={11121--11129},
  year={2023}
}

@article{zaidi2024unlocking,
  title={Unlocking the potential of digital twins in supply chains: A systematic review},
  author={Zaidi, Syed Adeel Haneef and Khan, Sharfuddin Ahmed and Chaabane, Amin},
  journal={Supply Chain Analytics},
  volume={7},
  pages={100075},
  year={2024},
  publisher={Elsevier}
}

@article{roman2025state,
 title={State of the Art of Digital Twins in Improving Supply Chain Resilience.},
  author={Roman, Eugenia-Alina and Stere, Armand-Serban and Roșca, Eugen and Radu, Adriana-Valentina and Codroiu, Denis and Anamaria, Ilie},
  journal={Logistics (2305-6290)},
  volume={9},
  number={1},
  year={2025}
}

@incollection{lee2025ai,
  title={AI-Driven Supply Chain Transformation},
  author={Lee, Hau L and Chen, Zhe and Qi, Yongzhi and Shen, Zuo-Jun Max},
  booktitle={AI in Supply Chains: Perspectives from Global Thought Leaders},
  pages={223--244},
  year={2025},
  publisher={Springer}
}

@article{cannas2024artificial,
  title={Artificial intelligence in supply chain and operations management: a multiple case study research},
  author={Cannas, Violetta Giada and Ciano, Maria Pia and Saltalamacchia, Mattia and Secchi, Raffaele},
  journal={International journal of production research},
  volume={62},
  number={9},
  pages={3333--3360},
  year={2024},
  publisher={Taylor \& Francis}
}

@article{donti2024taskbased,
  title={Task-based end-to-end model learning in stochastic optimization},
  author={Donti, Priya and Amos, Brandon and Kolter, J Zico},
  journal={Advances in neural information processing systems},
  volume={30},
  year={2017}
}

@article{bertsimas2020prescriptive,
  title={From predictive to prescriptive analytics},
  author={Bertsimas, Dimitris and Kallus, Nathan},
  journal={Management Science},
  volume={66},
  number={3},
  pages={1025--1044},
  year={2020},
  publisher={INFORMS}
}

@article{sadana2025survey,
  title={A survey of contextual optimization methods for decision-making under uncertainty},
  author={Sadana, Utsav and Chenreddy, Abhilash and Delage, Erick and Forel, Alexandre and Frejinger, Emma and Vidal, Thibaut},
  journal={European Journal of Operational Research},
  volume={320},
  number={2},
  pages={271--289},
  year={2025},
  publisher={Elsevier}
}

@inproceedings{zhou2021informer,
  title={Informer: Beyond efficient transformer for long sequence time-series forecasting},
  author={Zhou, Haoyi and Zhang, Shanghang and Peng, Jieqi and Zhang, Shuai and Li, Jianxin and Xiong, Hui and Zhang, Wancai},
  booktitle={Proceedings of the AAAI conference on artificial intelligence},
  volume={35},
  number={12},
  pages={11106--11115},
  year={2021}
}

@inproceedings{nie2023patchtst,
  title={A Time Series is Worth 64Words: Long-term Forecasting with Transformers},
  author={Nie, Y},
  journal={arXiv preprint arXiv:2211.14730},
  year={2022}
}

@inproceedings{wu2023timesnet,
  title={Timesnet: Temporal 2d-variation modeling for general time series analysis},
  author={Wu, Haixu and Hu, Tengge and Liu, Yong and Zhou, Hang and Wang, Jianmin and Long, Mingsheng},
  journal={arXiv preprint arXiv:2210.02186},
  year={2022}
}

@article{schmid2025comparing,
  title={Comparing statistical and machine learning methods for time series forecasting in data-driven logistics—A simulation study},
  author={Schmid, Lena and Roidl, Moritz and Kirchheim, Alice and Pauly, Markus},
  journal={Entropy},
  volume={27},
  number={1},
  pages={25},
  year={2024}
}

@article{gijsbrechts2022inventory,
  title={Can deep reinforcement learning improve inventory management? Performance on lost sales, dual-sourcing, and multi-echelon problems},
  author={Gijsbrechts, Joren and Boute, Robert N and Van Mieghem, Jan A and Zhang, Dennis J},
  journal={Manufacturing \& Service Operations Management},
  volume={24},
  number={3},
  pages={1349--1368},
  year={2022},
  publisher={INFORMS}
}

@article{qi2023e2e_inventory,
  title={A practical end-to-end inventory management model with deep learning},
  author={Qi, Meng and Shi, Yuanyuan and Qi, Yongzhi and Ma, Chenxin and Yuan, Rong and Wu, Di and Shen, Zuo-Jun},
  journal={Management Science},
  volume={69},
  number={2},
  pages={759--773},
  year={2023},
  publisher={INFORMS}
}





\end{document}